\documentclass[sigconf]{acmart}
\usepackage{graphicx}
\usepackage{amsmath}

\usepackage{amssymb}

\usepackage{xcolor}
\usepackage{colortbl}
\usepackage{bbding}
\usepackage{xfrac}
\usepackage{multirow}
\usepackage{balance}
\usepackage{pifont}
\usepackage{utfsym}
\AtBeginDocument{%
  }

\copyrightyear{2023} 
\acmYear{2023} 
\setcopyright{acmlicensed}\acmConference[MM '23]{Proceedings of the 31st ACM International Conference on Multimedia}{October 29-November 3, 2023}{Ottawa, ON, Canada}
\acmBooktitle{Proceedings of the 31st ACM International Conference on Multimedia (MM '23), October 29-November 3, 2023, Ottawa, ON, Canada}
\acmPrice{15.00}
\acmDOI{10.1145/3581783.3611715}
\acmISBN{979-8-4007-0108-5/23/10}

\begin{document}

\title{Exploring Shape Embedding for Cloth-Changing Person Re-Identification via 2D-3D Correspondences}

\author{Yubin Wang}
\orcid{0009-0009-6149-8707}
\affiliation{%
  \institution{College of Information Science and Electronic Engineering, Zhejiang University}
  \city{Hangzhou}
  \country{China}
  \postcode{310000}
}
\email{zjuwyb1999@gmail.com}

\author{Huimin Yu}
\orcid{0000-0002-8206-3747}
\authornote{Corresponding Author.}
\affiliation{%
  \institution{College of Information Science and Electronic Engineering, Zhejiang University}
  \city{Hangzhou}
  \country{China}
  \postcode{310000}
}
\email{yhm2005@zju.edu.cn}

\author{Yuming Yan}
\orcid{0009-0005-4596-1600}
\affiliation{%
  \institution{College of Information Science and Electronic Engineering, Zhejiang University}
  \city{Hangzhou}
  \country{China}
  \postcode{310000}
}
\email{12231016@zju.edu.cn}

\author{Shuyi Song}
\orcid{0009-0004-4524-1155}
\affiliation{%
  \institution{College of Information Science and Electronic Engineering, Zhejiang University}
  \city{Hangzhou}
  \country{China}
  \postcode{310000}
}
\email{22131091@zju.edu.cn}

\author{Biyang Liu}
\orcid{0000-0002-6224-2748}
\affiliation{%
  \institution{College of Information Science and Electronic Engineering, Zhejiang University}
  \city{Hangzhou}
  \country{China}
  \postcode{310000}
}
\email{11831033@zju.edu.cn}

\author{Yichong Lu}
\orcid{0009-0007-3246-0166}
\affiliation{%
  \institution{College of Information Science and Electronic Engineering, Zhejiang University}
  \city{Hangzhou}
  \country{China}
  \postcode{310000}
}
\email{luyi200106@gmail.com}

\renewcommand{\shortauthors}{Yubin Wang et al.}

\begin{abstract}
Cloth-Changing Person Re-Identification (CC-ReID) is a common and realistic problem since fashion constantly changes over time and people's aesthetic preferences are not set in stone. While most existing cloth-changing ReID methods focus on learning cloth-agnostic identity representations from coarse semantic cues (e.g. silhouettes and part segmentation maps), they neglect the continuous shape distributions at the pixel level. In this paper, we propose Continuous Surface Correspondence Learning (CSCL), a new shape embedding paradigm for cloth-changing ReID. CSCL establishes continuous correspondences between a 2D image plane and a canonical 3D body surface via pixel-to-vertex classification, which naturally aligns a person image to the surface of a 3D human model and simultaneously obtains pixel-wise surface embeddings. We further extract fine-grained shape features from the learned surface embeddings and then integrate them with global RGB features via a carefully designed cross-modality fusion module. The shape embedding paradigm based on 2D-3D correspondences remarkably enhances the model's global understanding of human body shape. To promote the study of ReID under clothing change, we construct 3D Dense Persons (DP3D), which is the first large-scale cloth-changing ReID dataset that provides densely annotated 2D-3D correspondences and a precise 3D mesh for each person image, while containing diverse cloth-changing cases over all four seasons. Experiments on both cloth-changing and cloth-consistent ReID benchmarks validate the effectiveness of our method. Our project page is located at \url{https://CSCL-CC.github.io}.
\end{abstract}


\maketitle
\begin{figure}[!t]
    \centering
    \includegraphics[width=0.48\textwidth]{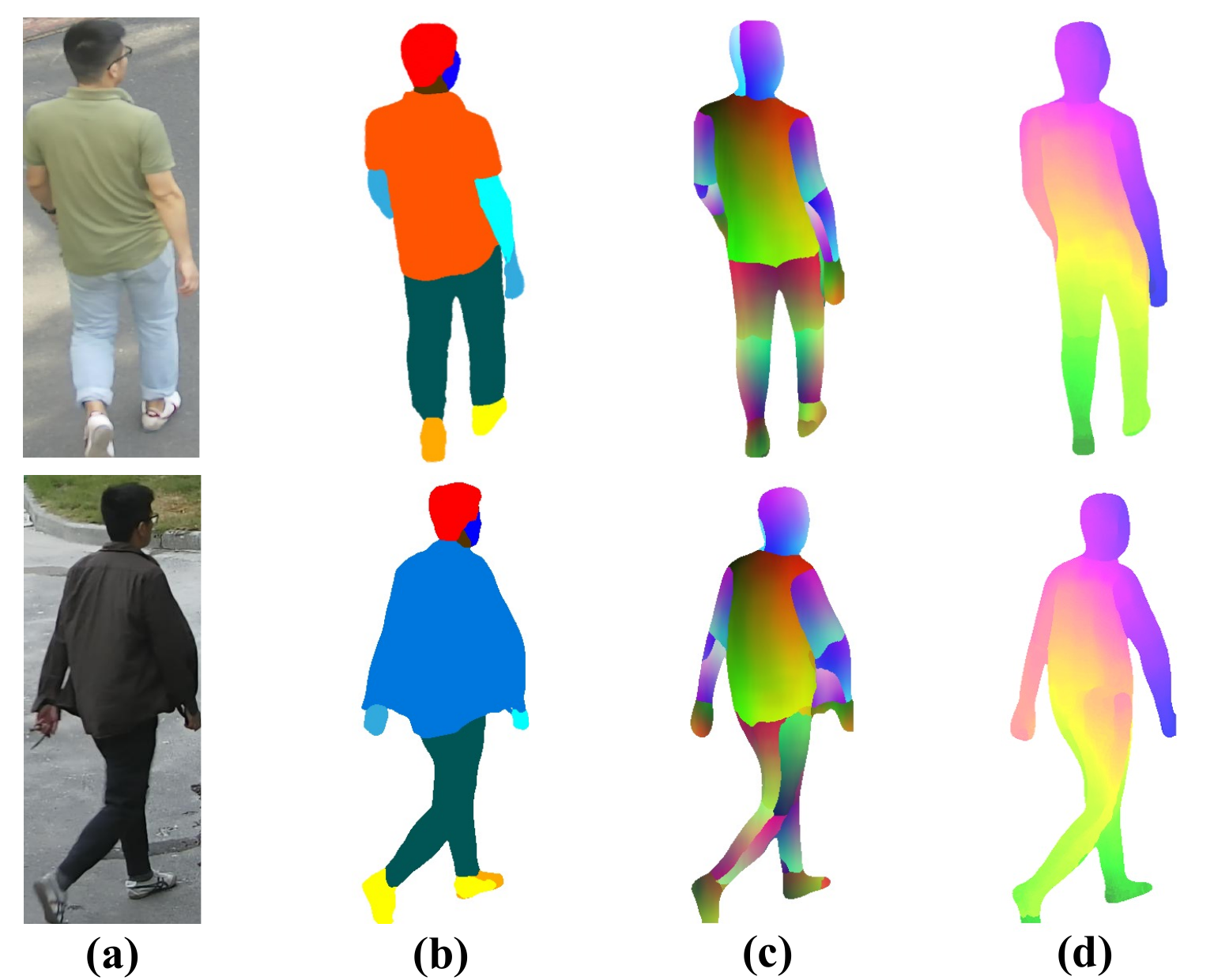}
    \caption{Comparison of different multi-modal auxiliary information for person re-identification. (a) Images of the same person in DP3D; (b) Coarse part segmentation, with only part labels estimated; (c) Discretized DensePose IUV estimation, with obvious seams between body parts; (d) Continuous 2D-3D correspondences between image pixels and the entire body surface, obtained through our CSCL framework.}
    \label{fig:figure1}
    \vspace{-0.4cm}
\end{figure}
\section{Introduction}
\label{sec:intro}
%
Person Re-Identification (Re-ID) targets at re-identifying a specific person across disjoint cameras~\cite{surv2022}. Most existing works ~\cite{GeneralPerson1, VAF, GeneralPerson4, General5, General6, General7, WACV2023} presuppose that the appearances of people remain consistent over time. In reality, people tend to change their outfits over a long duration and different people may share the same dressing sense. Methods that rely excessively on clothing appearance fail to generalize to this long-term cloth-changing scenario. 

In recent years, plenty of efforts~\cite{Distengled2023, Adversarial2021,Win-Win, SemanticGuided2021, SpeRe-Ranking, Patching2022, GaitCC} have been made to handle the cloth-changing issue by learning discriminative cloth-agnostic identity representations.  A small proportion of methods~\cite{Adversarial2021,Distengled2023, Win-Win} attempt to decouple cloth-agnostic features directly from RGB images without multi-modal auxiliary information, which inevitably leads to the loss of crucial information in global features and results in a heavy reliance on the domain. The mainstream methods~\cite{SemanticGuided2021, SpeRe-Ranking, Patching2022, IRANet2022, multigranular2022, GaitCC} typically adopt human parsing models to obtain coarse semantic cues to guide the extraction of biometric features, such as shape features. However, as shown in Figure~\ref{fig:figure1}(b), coarse semantic cues are insufficient to obtain detailed shape information of a specific person, as it only enables the estimation of body part labels but fails to model pixel-wise shape distributions within the parts. Several recent works~\cite{Densely2019, AlignTransformer} leverage dense pose estimation~\cite{DensePose2018} to align the texture of body parts based on UV mapping. However, they do not further explore reliable shape representations for the ReID task. Additionally, these methods have a major defect in that they require partitioning the 3D model into charts, and the resulting discretized UV spaces prevent them from learning continuous correspondences over the entire body surface. As shown in Figure~\ref{fig:figure1}(c), the use of independent UV coordinate systems for each body part results in noticeable part seams in the estimated IUV maps. There are also some methods~\cite{3DSL, CoAttCC} directly estimating SMPL~\cite{smpl2015} shape parameters as 3D shape features. However, the SMPL shape parameter space is highly incompatible with the image feature space, making it challenging to effectively integrate features from these two modalities.

In this paper, we propose a Continuous Surface Correspondence Learning (CSCL) framework, which represents a new shape embedding paradigm for cloth-changing ReID. CSCL pixel-wisely maps a person image to a continuous embedding space of the SMPL mesh surface through vertex classification. Essentially, learning continuous 2D-3D correspondences aligns a person image to the entire surface of a 3D human model, and simultaneously obtains a pixel-level continuous distribution of body shape on the canonical 3D surface. Even for different persons wearing the same clothes, there can be significant differences in their body shape distributions. Therefore, we further extract fine-grained discriminative shape features from the established correspondences, and integrate them with global RGB features via an optimized cross-modality fusion module based on the transformer~\cite{Transformer}, which greatly compensates for the lost shape details in global RGB features. We incorporate a novel Latent Convolutional Projection (LCP) layer for feature projection. The LCP layer enhances the sharing and correlation among tokens via adding an additional latent embedding, which is the latent vector of an auto-encoder designed to reconstruct the token map. It is also noteworthy that the proposed framework generalizes well to the cloth-consistent cases, indicating the reliability of the learned shape features.

However, there is currently no publicly available cloth-changing ReID dataset with ground-truth dense 2D-3D correspondences. To facilitate the research, we construct a large-scale cloth-changing ReID dataset named 3D Dense Persons (DP3D), which contains 39,100 person images of 413 different persons captured by 15 cameras over all four seasons. We annotated dense 2D-3D correspondences for each person image via a carefully designed annotation system, ensuring 80 to 125 annotations for each image. 
\par The main contributions of this work are summarized as follows:
\begin{itemize}
    \item We propose a new shape embedding paradigm for cloth-changing ReID that establishes pixel-wise and continuous correspondences between a 2D image plane and a canonical 3D human body surface. To the best of our knowledge, this is also the first work to explore global shape representations for cloth-changing ReID via 2D-3D correspondences.
    \item We develop an optimized cross-modality fusion module to adaptively integrate shape features with global RGB features, where a novel Latent Convolutional Projection (LCP) layer is designed to perform feature projection.
    \item We construct 3D Dense Persons (DP3D), which is the first large-scale cloth-changing ReID dataset with densely annotated 2D-3D correspondences and a corresponding 3D mesh for each person image, while containing highly diverse cloth-changing cases in real-world scenarios. 
    \item We demonstrate our proposed method is applicable to both cloth-changing and cloth-consistent situations, as shown by extensive results on four cloth-changing ReID datasets including DP3D and two general ReID datasets.
\end{itemize}

\section{Related Works}
\label{sec:related_works}
In this section, we first review the literature on cloth-changing person re-identification and corresponding datasets, then introducing the research related to continuous surface embeddings in the context of 3D shape analysis.
\subsection{Cloth-Changing Person ReID}
Existing cloth-changing ReID methods can be categorized into decoupling-based methods and auxiliary modality-based methods. Decoupling-based methods~\cite{Adversarial2021, Weakening2022, RGBOnly2022} aim to decouple cloth-agnostic features directly from RGB images without multi-modal auxiliary information. AFD-Net~\cite{Adversarial2021} disentangled identity and clothing features via generative adversarial learning. CAL~\cite{RGBOnly2022} proposed to penalize the predictive power of the ReID model with respect to clothes via a clothes-based adversarial loss, while UCAD~\cite{Weakening2022} enforced the identity and clothing features to be linearly independent in the feature space via an orthogonal loss.

Auxiliary modality-based methods~\cite{SemanticGuided2021, multigranular2022, ShapeAppearance2021, 3DSL, Densely2019, AlignTransformer} are considered more robust since visual texture features can be filtered under the supervision of human semantics. FSAM~\cite{ShapeAppearance2021} proposed to complement 2D shape representations obtained from human silhouettes for global features. MVSE~\cite{multigranular2022} embedded multigranular visual semantic information into the model. Pixel Sampling~\cite{SemanticGuided2021} leveraged a human parsing model to recognize upper clothes and pants, and then randomly changed them by sampling pixels from other people, enforcing the model to automatically learn cloth-agnostic cues. DSA-ReID\cite{Densely2019} and ASAG-Net\cite{AlignTransformer} proposed to use dense human semantics to generate semantics-aligned images in the discretized DensePose UV space, while 3DSL~\cite{3DSL} considered the low-dimensional SMPL shape parameters as 3D shape features, and directly fused them to global features. None of these methods consider establishing pixel-wise and continuous 2D-3D correspondences between image pixels and the entire 3D body surface, which effectively bridges the gap between 2D and 3D shape space.
\subsection{Cloth-Changing ReID Datasets}
General person ReID datasets\cite{Market2015, CUKE03, DukeMTMC, MSMT172017} assume that the appearance of the same individual is consistent, which is often not the case in real-world scenarios. Models trained on these datasets rely excessively on clothing appearance, making it difficult for them to generalize well to long-term cloth-changing scenarios. In recent years, a few datasets were collected specifically for the cloth-changing setting. Celebrities~\cite{Celeb2019} were obtained from the Internet, which consists of street snapshots of celebrities. PRCC~\cite{PRCC} provides indoor cloth-changing person images with their corresponding contour sketches. COCAS~\cite{COCAS} is a large-scale dataset that provides a variety of clothes templates for cloth-changing person ReID. LTCC~\cite{LTCC2020} assumes that different people wear different clothes and assigns a unique clothing label to each person image in the dataset. VC-Clothes~\cite{VCClothes} is a large realistic synthetic dataset rendered by the GTA5 game engine. CSCC~\cite{Weakening2022} considers different degrees of cloth-changing. NKUP~\cite{NKUP} contains both indoor and outdoor person images with complex illumination conditions, while NKUP+~\cite{NKUP+} has more diverse scenarios, perspectives, and appearances.

\begin{table}[!t]
\small
\caption{Comparison of DP3D and existing cloth-changing ReUD datasets (`In': Indoor; `Out': Outdoor).}
\centering
\renewcommand{\arraystretch}{1.15}
\resizebox{0.485\textwidth}{!}{
\begin{tabular}{c|ccccccc}
\hline
Datasets &Scene&IDs& Image & Cam & Time& 3D View & Dense Corr.\\
\hline
Celebrities \cite{Celeb2019} &-&590&10,842&-&-&\usym{2717}&\usym{2717}\\
LTCC \cite{LTCC2020} &In& 152&17,138&12&2 Months&\usym{2717}&\usym{2717} \\
PRCC \cite{PRCC} &In& 221&33,698&3&-&\usym{2717}&\usym{2717} \\
COCAS \cite{COCAS} &In& 5,266&62,382&30&-&\usym{2717}&\usym{2717}\\
VC-Clothes \cite{VCClothes} &-&512&19,060&-&-&\usym{2717}&\usym{2717}\\
CSCC \cite{Weakening2022}&Out& 267 &36,700&13&12 Months&\usym{2717}&\usym{2717}\\
NKUP \cite{NKUP} &In/Out& 107 &9,738&15&4 Month&\usym{2717}&\usym{2717}\\
NKUP+ \cite{NKUP+} &In/Out& 361 &40,217&29&10 Month&\usym{2717}&\usym{2717}\\
DP3D (Ours) &Out& 413 &39,100&15&12 Months&\usym{2713}&\usym{2713}\\
\hline
\end{tabular}
}
\label{tab:dataset_compare}
\vspace{-0.5cm}
\end{table}
\subsection{Continuous Surface Embeddings}
Continuous Surface Embeddings (CSE) target at pixel-wisely learning an embedding of the corresponding 3D vertex from an RGB image~\cite{Continuous2020}, which demonstrates strong human body representation capabilities. HumanGPS~\cite{HumanGPS2021} employs contrastive learning to enhance CSE representations. BodyMap~\cite{BodyMap2022} introduced a coarse-to-fine learning scheme, establishing high-definition full-body continuous correspondences by refining coarse correspondences. SurfEmb~\cite{SurfEmb} applied Continuous Surface Embeddings to the field of object pose estimation and learned correspondence distributions in a self-supervised fashion. 

\begin{figure}[!t]
    \centering
    \includegraphics[width=0.485\textwidth]{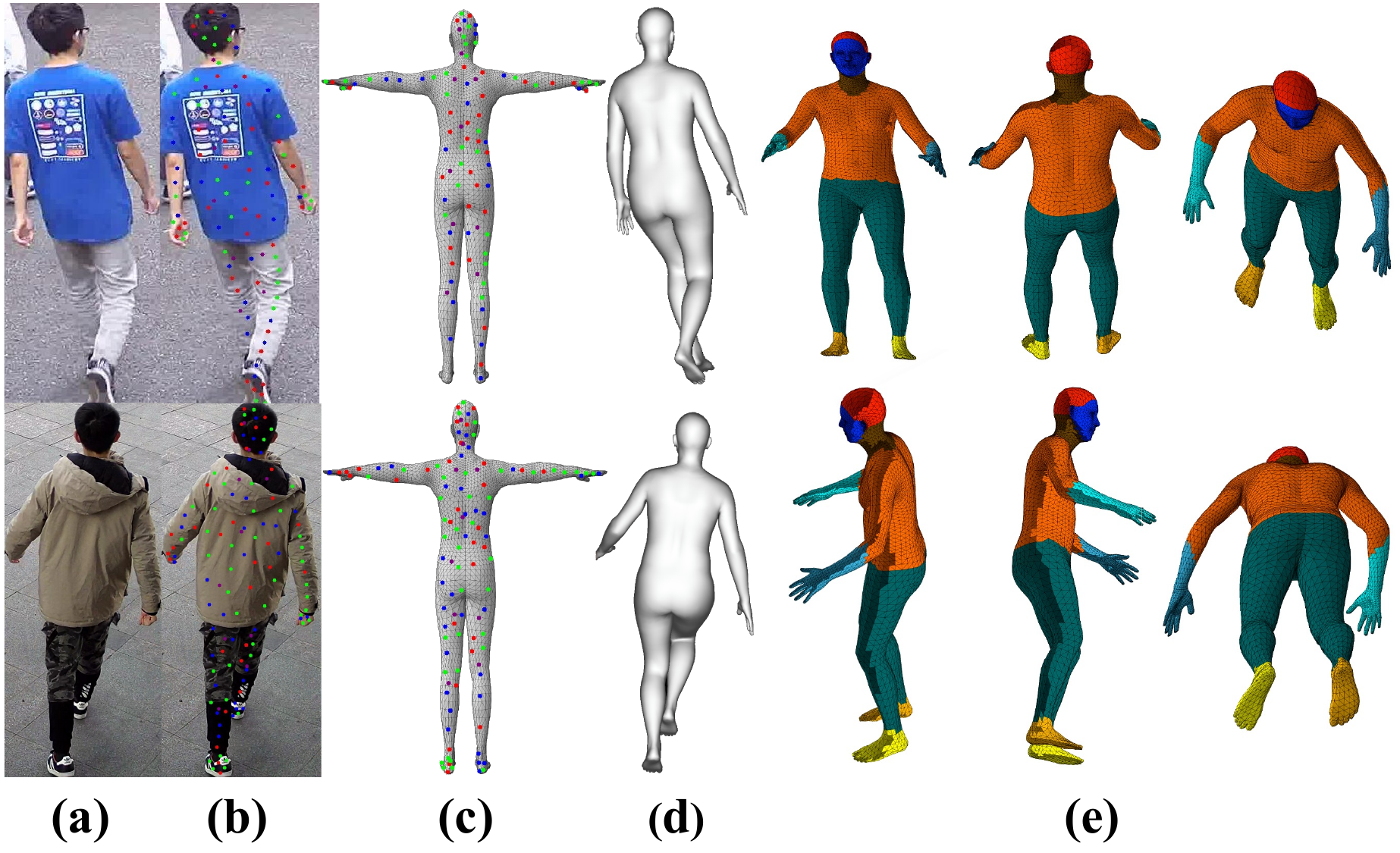}
    \caption{Examples of annotating person images in the DP3D dataset. (a) Cross-appearance images of the same person; (b) Generating pixels to be labeled (corresponding pixels are visualized with purple dots); (c) Annotating ground-truth corresponding 3D mesh vertices. (d) fitting the SMPL model to the person images under the guidance of dense correspondences; (e) the projected 2D full-body images used for annotation.}
    \label{fig:anno}
    \vspace{-0.2cm}
\end{figure}
\begin{figure*}
    \centering
    \includegraphics[width=\textwidth]{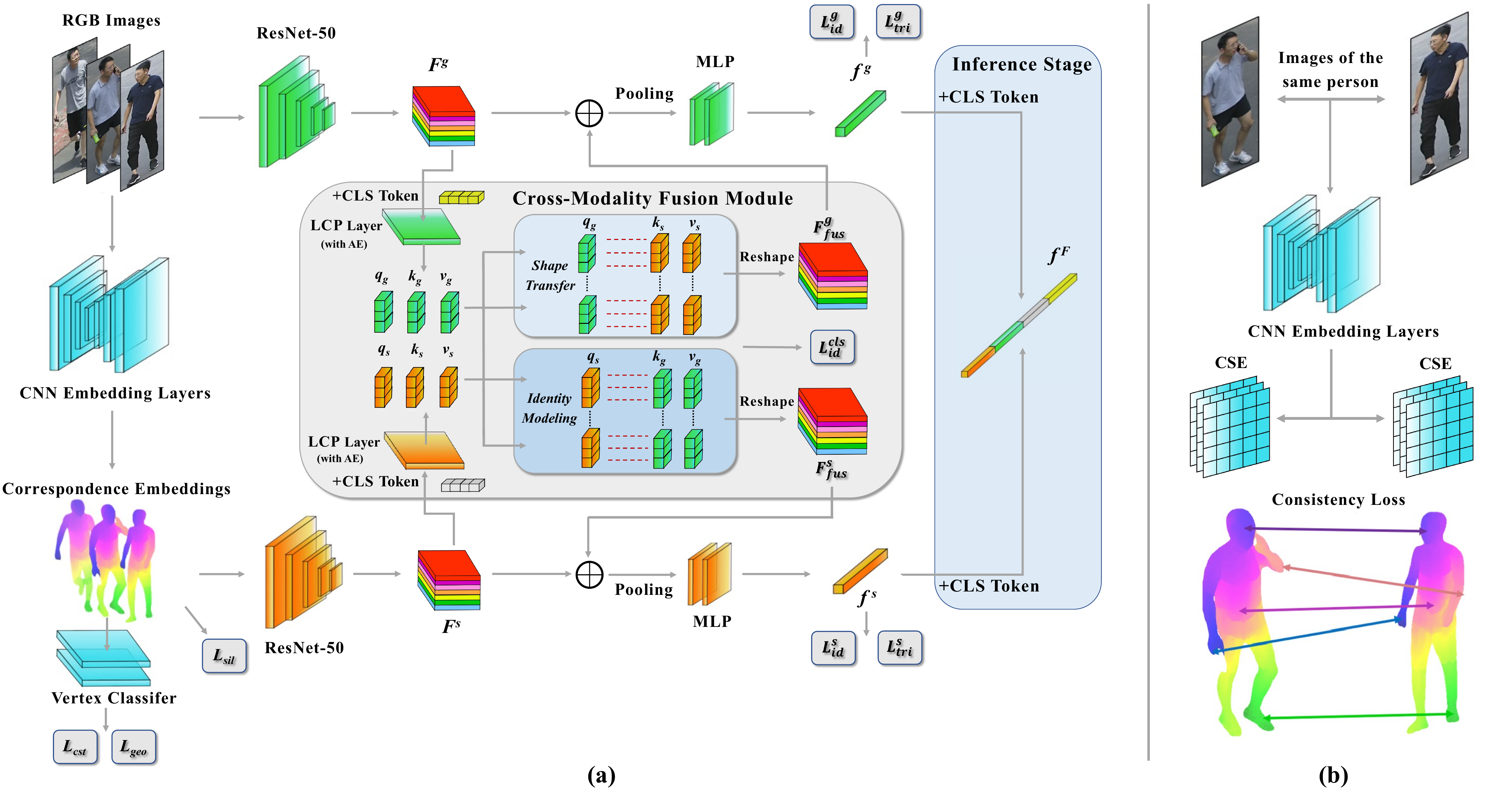}
    \caption{The architecture of the CSCL framework. (a) Our framework learns pixel-wise and continuous 2D-3D correspondences, which enables the extraction of fine-grained shape features. Cloth-agnostic shape knowledge is then complemented for global RGB features via cross-modality fusion; (b) Consistency learning between cross-view corresponding pixels.} 
    \label{fig:model1}
\end{figure*}
\section{the 3D Dense Persons Dataset}
Obtaining ground-truth 3D structure information for pedestrians is of substantial importance as it can address potential geometric ambiguities that may arise from relying solely on RGB modality. 

In this section, we introduce the 3D Dense Persons (DP3D), a large-scale cloth-changing ReID dataset that provides densely annotated 2D-3D correspondences and a corresponding 3D mesh for each person image, filling the gap in the field. 

\subsection{Data Collection}
 The raw videos we collected have high resolutions and cover a time span of one year. We selected a total of 15 cameras, with 5 of them having a resolution of 4K, 2 having a resolution of 2K, and the remainder being set to a resolution of 1080P. The use of high-resolution cameras ensures the recorded pedestrians to be as clear as possible, which is advantageous for the ReID task under clothing change. The shooting scenes encompass various outdoor locations, such as street scenes, park landscapes, construction sites, and parking lots. All pedestrians were captured by at least 2 cameras, with the majority being captured by 3 or more. We adopted the Mask R-CNN~\cite{MaskRCNN2017} framework to detect the bounding box of each person after framing. 

\subsection{Annotation System}
Due to the dramatic variations in people's clothing styles over the course of a year, we first identified the volunteers and conducted a manual inspection to avoid misidentification, while assigning a camera ID label, a person ID label, and a clothing ID label to each person image. Then, as shown in Figure~\ref{fig:anno}, we annotated dense correspondences via a carefully designed pipeline. In the first stage, we ran the universal model of Graphonomy~\cite{Graphonomy} with 20 part labels to segment the images, then uniformly sampling 40 pixels across the entire human body region. We also utilized k-means clustering to obtain 5 to 10 centroid pixels for each part based on its size. Compared to DensePose~\cite{DensePose2018}, our sampling method avoids seams between body parts and ensures a sufficient number of sampling points for smaller parts. However, since people may wear loose clothes, we manually filtered out those sampling pixels that did not fall within the human body regions underneath the clothes. For each pair of images belonging to the same person, we additionally selected 10 corresponding pixels for consistency learning, which correspond to the same 10 mesh vertices. In the second stage, as shown in Figure~\ref{fig:anno} (e), we projected the SMPL mean template mesh from 6 predefined viewpoints to generate full-body images. When annotating a specific pixel, it was only necessary to choose the most suitable projected image, and its 2D coordinates were used to localize the corresponding 3D vertex. In cases certain pixels were challenging to determine from the projected images, we directly annotated the correspondences on the 3D mesh surface through rotation. It is worth noting that we did not annotate in a part-by-part manner, but rather adopted a global approach using full-body projected images for annotation, which ensured accurate annotations at the junctions of body parts. In the last stage, to obtain accurate SMPL parameters, we employed a modified SMPLity-X~\cite{Smplify-X} to fit the SMPL model to the person images under the guidance of densely annotated correspondences. 
\subsection{Statistics and Comparison}
The proposed DP3D dataset is characterized by its diverse scenes, multiple perspectives, large number of individuals, and long time span. It comprises 39,100 person images belonging to 413 different persons, which were captured over the course of a year (during four distinct seasons). Depending on its resolution, each person image has approximately 80 to 125 annotated correspondences, where 10 correspondences have mesh vertices shared among all images of the same person. We divided the images into a training set and a testing set, with each set containing approximately equal numbers of identities. For same-appearance images of a specific person, we randomly select one image per viewpoint to construct the query set, while the remaining images in the testing set form the gallery set. We present in Table~\ref{tab:dataset_compare} a comparison between DP3D and existing cloth-changing ReID datasets. 
\section{Methodology}
\label{sec:approach}
In this section, we first provide an overview of our proposed framework in Section~\ref{subsec:overview}. Next, in Section~\ref{subsec:cse} and ~\ref{subsec:fusion}, we elaborate the learning scheme of continuous 2D-3D correspondences, as well as the design principles of the cross-modality fusion module, respectively. Subsequently, we provide a comprehensive description of the training losses in Section~\ref{subsec:loss}.
\subsection{Overview}
\label{subsec:overview}
As shown in Figure~\ref{fig:model1} (a), person images are input separately into the ResNet-50~\cite{HeRes} backbone and CNN embedding layers to extract global RGB features and continuous surface embeddings. For each foreground pixel, CSCL maps it to a continuous embedding space of the SMPL mesh surface under the supervision of geodesic distances. Subsequently, a shape extraction network with a ResNet-50 architecture is further employed to extract fine-grained shape features from the learned surface embeddings, while simultaneously mapping them to the same size as global RGB features. Following that, we adaptively integrate shape features with global RGB features via an improved cross-modality fusion module, where a novel Latent Convolutional Projection (LCP) layer is designed to perform feature projection. Cross-attention mechanism is then applied to aggregate features from the two distinct modalities, which are then added to the original features. After the fusion, we conduct Global Average Pooling (GAP), followed by two separate fully-connected classifiers, to obtain the final global RGB features and shape features. We also introduce a learnable class token for each of the two modalities, which exhibits strong cross-modality compatibility and also contributes to the ID loss. In the inference stage, the two class tokens are concatenated with global RGB features and shape features to construct the final identity feature.
\subsection{Establishing Continuous Correspondences}
\label{subsec:cse}
Considering the huge domain gap between 2D person images and the 3D space perceived by human eyes, we believe that establishing continuous correspondences between image pixels and the entire 3D human body is of substantial importance, which bridges the gap between the 2D and 3D shape space and therefore benefit the understanding of global body shape.

Given a person image $I\in\mathbb{R}^{H\times W\times 3}$ of height $H$ and width $W$, we first extract the segmentation mask $M$ of the foreground person. Then, the CNN embedding layers map the person image into continuous surface embeddings $E\in\mathbb{R}^{H\times W\times D}$, while preserving the spatial resolution of the image. For pixels within the foreground mask $M$, we employ geodesic distances on the 3D surface to supervise the learning of surface embeddings. More concretely, we scale the cross-entropy loss of pixel-to-vertex classification on the mesh surface using geodesic distances. This constraint is reasonable as it quantifies the deviation of vertex prediction on the 3D surface. Furthermore, as illustrated in Figure~\ref{fig:model1} (b), we also conduct consistency learning for corresponding pixels in images that belong to the same person. Suppose we have two distinct images of the same person, donated as $I_1$, $I_2$, where foreground pixels $p_1$ and $p_2$ belong to image $I_1$, and pixel $q$ belongs
to image $I_2$. Both $p_1$ and $q$ correspond to the same vertex $v_1$ on the mesh surface, while $p_2$ corresponds to vertex $v_2$. We first compute the cosine distance in the embedding space to measure the similarity between $p_1$ and $q$:
\begin{equation}
    d(p_1, q) = 1 - cos(E_1(p_1), E_2(q))
\end{equation}
where $E_1$ and $E_2$ denote surface embeddings of images $I_1$ and $I_2$.
By minimizing the cosine distance $d(p_1, q)$, the embedding vectors of two corresponding pixels are brought closer. However, during training, only considering the consistency of corresponding pixels may lead to all embeddings mapping to similar values. Therefore, for different pixels $p_1$ and $p_2$ in the same person image, we keep their relative affinity by enforcing embedding distances to follow geodesic distances, i.e. minimizing $\lvert d(p_1, p_2) - s(g(v_1, v_2))\rvert$, where $g(\cdot, \cdot)$ calculates the geodesic distance between two mesh vertices and $s(\cdot)$ scales it to match the range of the cosine distance $d(\cdot, \cdot)$.

Establishing 2D-3D correspondences allows for learning the continuous shape distributions on the 3D surface at the pixel level, i.e. $Pr(v|I, p, p\in M)$, where I denotes the person image, and M denotes the foreground mask. To further extract fine-grained shape features, we feed the learned embeddings into the shape extraction network with a ResNet-50 architecture, while mapping them to the same size as global RGB features. Note that the extracted shape features are insensitive to clothing appearance as texture features are already filtered out in the correspondence learning process.
\subsection{Cross-Modality Feature Fusion}
\label{subsec:fusion}
To adaptively integrate the shape features extracted from the established continuous correspondences with global RGB features, a cross-modality fusion module is designed. As discussed in CVT~\cite{CVT}, convolutional layers are renowned for their remarkable ability to capture intricate local spatial token structures, which allows the removal of positional embeddings from the transformer~\cite{Transformer} framework. However, the utilization of fixed-size convolutional kernels hampers the effectiveness of capturing global positional correlations between non-adjacent tokens. To mitigate this issue, we propose a novel Latent Convolutional Projection (LCP) layer. It adds the same latent embedding to each token in the token map, which is the latent vector of a pretrained auto-encoder designed to reconstruct the token map. During the training of CSCL, only the encoder of the auto-encoder is preserved and fixed to ensure the universal nature of the latent embedding, whereas the decoder is disregarded. This design not only greatly enhances the correlation and sharing among different tokens, but also enables better adaptation to images with diverse backgrounds. The projection of an LCP layer can be formulated as follows:
\begin{equation}
    Q/K/V = Flatten(Conv2d(Reshape2D(F) + l))
\end{equation}
where $Q/K/V$ represents the projected queries, keys, and values, $F$ is the input token map,  $l$ represents the latent embedding, and $Reshape2D$ denotes the operation to reshape the feature map $F$ to a 2D token map. After separately passing global RGB features $F^{g}\in \mathbb{R}^{h\times w\times c}$ and shape features $F^s\in\mathbb{R}^{h\times w\times c}$ through two distinct LCP layers, the cross-attention mechanism is applied to adaptively integrate features from different modalities. We first take global RGB features as queries and shape features as keys/values, reshape the fused feature to match the size of $F^{g}$, and finally add it to $F^{g}$:
\begin{equation}
    F^{g} = F^{g} + Reshape3D(MHA(Q_{g}, K_s, V_s))
\end{equation}
where $Reshape3D$ denotes the operation of reshaping a 2D token map to match the size of $F^{g}$, and MHA represents the multi-head attention.
We also take shape features as queries and global RGB features as keys/values for identity modeling of shape features. 
\begin{equation}
    F^{s} = F^{s} + Reshape3D(MHA(Q_{s}, K_g, V_g))
\end{equation}
In other words, we enable bidirectional access between global RGB features and shape features, which allows the model not only complements fine-grained cloth-agnostic shape knowledge for global RGB features $F^{g}$, but also  integrates essential identity-related characteristics for shape features $F^{s}$ to assist identity modeling. Additionally, we introduce learnable class tokens for each of the two modalities, which are also utilized to compute the ID loss. 
\subsection{Loss Function}
\label{subsec:loss}
\textbf{CSE Losses.} As discussed in Section~\ref{subsec:cse}, to mask out the background pixels, the foreground silhouette for each person image is retrieved, thus a binary cross-entropy loss $\mathcal{L}_{sil}$ is employed to penalize unsatisfactory silhouette predictions. Furthermore, we employ geodesic distances on the mesh surface to scale the per-pixel vertex classification loss, which penalizes the misclassified pixels based on the degree of deviation on the surface. The geodesic loss can be formulated as follows:
\begin{equation}
    \mathcal{L}_{geo} = -\frac{1}{N}\sum_{p\in I} {g(v_{p}, \hat{v_p})\cdot log(p(\hat{v_p}))}
\end{equation}
where $N$ indicates the number of pixels with ground-truth annotations in image $I$, $v_p$ and $\hat{v_p}$ represent the ground-truth and predicted mesh vertices corresponding to pixel p, and $g(\cdot, \cdot)$ calculates geodesic distances between two mesh vertices. For consistency learning of continuous surface embeddings, we design the following consistency loss $\mathcal{L}_{cst}$: 
\begin{equation}
\begin{aligned}
    \mathcal{L}_{cst} &= \frac{1}{N_1}\sum_{p\in I_1, q\in I_2} {log(1+exp(d(p, q))} \\&+ \frac{1}{N_2}\sum_{p_1, p_2\in I} {log(1+exp(\lvert d(p_1, p_2) -  s(g(v_1, v_2))\rvert))}
\end{aligned}
\end{equation}
where $N_1$ and $N_2$ indicate the number of annotated pairs, $p$ and $q$ are corresponding pixels in cross-view images, $p_1$ and $p_2$ stand for different pixels in the same image, $d(\cdot, \cdot)$ and $g(\cdot, \cdot)$ respectively denote the cosine distance in the embedding space and the geodesic distance on the surface, and $s(\cdot)$ represents the scale function. The first term of $\mathcal{L}_{cst}$ ensures consistency between embeddings of cross-view corresponding pixels, while the second term enforces embedding distances to follow geodesic distances for different pixels in the same image, thus pushing apart their embeddings, and avoiding the degradation cases that may occur during training.

\textbf{ReID Losses.} The ReID losses employed in our framework consist of a cross-entropy loss (ID loss) for classification and a triplet loss~\cite{triplet} for similarity learning in the feature space. The final global RGB feature $f^{g}$, shape feature $f^{s}$, and two class tokens all contribute to the ID loss:
\begin{equation}
    \mathcal{L}_{id} = \mathcal{L}_{id}^g + \mathcal{L}_{id}^s + \mathcal{L}_{id}^{cls}
\end{equation}
where $\mathcal{L}_{id}^{cls}$ represents the summation of ID losses of the two class tokens. We introduce separate triplet losses for global RGB features and shape features to enhance their discriminative capability, which are combined to obtain the final triplet loss:
\begin{equation}
    \mathcal{L}_{tri} = \mathcal{L}_{tri}^g + \mathcal{L}_{tri}^s
\end{equation}

\textbf{Final Loss.} The overall objective function of our proposed Continuous Surface Correspondence Learning (CSCL) framework compromises the aforementioned CSE losses and ReID losses, which can be formulated as follows:
\begin{equation}
    \mathcal{L} = \mathcal{L}_{sil} + \lambda_1(\mathcal{L}_{geo} + \alpha\mathcal{L}_{cst}) + \lambda_2\mathcal{L}_{id} + \lambda_3\mathcal{L}_{tri}
    \label{eq:floss}
\end{equation}
where $\lambda_1$, $\alpha$, $\lambda_2$ and $\lambda_3$ are weights for balancing each term.

\begin{table*}[h]
\small
\caption{Comparison on LTCC, PRCC, VC-CLothes and DP3D datasets. \# denotes we conducted experiments based on the code we reproduced. `Standard',
`Cloth-Changing' and `Same-Clothes' represent experiment settings illustrated in Section~\ref{subsec:protocal}.}
\centering
\resizebox{1\textwidth}{!}{
\renewcommand{\arraystretch}{1.15}
\begin{tabular}{c|cc|cc|cc|cc|cc|cc|cc|cc}
\hline
\multirow{3}{*}{Methods} & \multicolumn{4}{c|}{LTCC} & \multicolumn{4}{c|}{PRCC}& \multicolumn{4}{c|}{VC-Clothes}& \multicolumn{4}{c}{DP3D} \\

\cline{2-5} \cline{6-7} \cline{8-11} \cline{12-17}
\multirow{1}{*}{} & \multicolumn{2}{c|}{Standard} & \multicolumn{2}{c|}{Cloth-Changing}&\multicolumn{2}{c|}{Same-Clothes} & \multicolumn{2}{c|}{Cloth-Changing}&\multicolumn{2}{c|}{Same-Clothes} & \multicolumn{2}{c|}{Cloth-Changing} & \multicolumn{2}{c|}{Standard}&\multicolumn{2}{c}{Cloth-Changing}\\
\cline{2-3} \cline{4-5} \cline{6-6}\cline{7-7} \cline{8-9}  \cline{10-11} \cline{12-17} 
& Rank-1      &  \multicolumn{1}{c|}{mAP} & Rank-1        &  \multicolumn{1}{c|}{mAP}  & Rank-1  &mAP    & Rank-1 &mAP   & Rank-1      &  \multicolumn{1}{c|}{mAP} & Rank-1        &  \multicolumn{1}{c|}{mAP} & Rank-1             &  mAP&Rank-1             &  mAP \\
\hline
PCB (ECCV18)~\cite{GeneralPerson3}       &    65.1     &    30.6   &    23.5    &    10.0    &86.9&83.6&22.9&24.7 &72.3&73.9&53.9&55.6&58.3&35.9&15.1&9.9\\
HACNN (CVPR18)~\cite{HACNN}    &    60.2     &    26.7    &    21.5    &    9.2      &82.4&84.7&21.8&23.2&68.6&69.7&49.6&50.1&53.4&31.8&13.4&8.5 \\
MGN (MM18)~\cite{MGN}  &    68.4     &    32.4    &    25.3    &    11.5     &89.8&87.4&25.9&35.9 &74.3&75.2&55.0&57.3&59.7&37.0&17.9&12.2\\
TransReID (ICCV21)~\cite{TransReID}  &    70.1     &    33.8    &    26.4    &    12.6     &93.1&94.0&40.1&43.6 &79.8&80.3&73.1&74.9&62.5&37.5&18.5&12.7\\
\hline
SE+CESD (ACCV20)~\cite{LTCC2020}   &    71.4     &    34.3    &    26.2    &    12.4    &    91.8 &90.6  &  37.6&38.7   &85.2  &79.1&69.5&65.5&61.9&38.3&18.3&12.7\\
FSAM (CVPR21)~\cite{ShapeAppearance2021}&73.2&35.4    &   38.5&16.2   &    98.8   &-&  54.5 &-  &94.7  &94.8&78.6&78.9&61.7&39.0&17.7&11.9\\
3DSL (CVPR21)~\cite{3DSL}&73.8\#&34.2\#    &   31.2&14.8   &    98.7\#   &95.0\#&  51.3 &49.8\#  &  92.5\#&79.7\#&79.9&81.2&66.4\#&45.3\#&29.6\#&17.8\#\\
UCAD (IJCAI22)~\cite{Weakening2022}     &    74.4    &    34.8   &    32.5    &    15.1&96.5&95.9&45.3&45.2 &92.6&81.1&82.4&73.8&63.5&41.7&21.3&13.1\\
MVSE (MM22)~\cite{multigranular2022}&73.4\#&33.9\#&\textbf{70.5}&33.0  &    98.7\#   &98.3\#&  47.4 &52.5  &86.1\#  &79.5\#&79.4\#&79.1\#&63.7\#&41.7\#&21.2\#&13.4\#\\
M2NET (MM22)~\cite{NKUP+}&-&-&-&-  &    99.5   &99.1&  59.3 &57.7  &-&-&-&-&63.3&39.4&20.8&12.9\\
CAL (CVPR22)~\cite{RGBOnly2022}&74.2&40.8    &   40.1&18.0   &    \textbf{100}   &\textbf{99.8}&  55.2 &55.8  &-  &-&-&-&64.8&42.4&22.9&14.4\\

\hline
Baseline(ResNet-50) &   68.2      &   34.3    &    26.2    &    12.3    &89.6& 88.0 &32.8&37.1&78.0&78.8&70.6&65.9&62.5&38.8&19.2&13.0\\
CSCL(w/o. $\mathcal{L}_{cst}$) &    75.3    &    41.1   &    68.9    &    33.5&99.7&99.4&63.5&63.6 &97.1&95.4&85.5&\textbf{84.7}&74.1&55.8&37.8&27.0\\
\textbf{CSCL}     &    \textbf{75.5}    &    \textbf{41.6}   &    69.7    &    \textbf{34.1}&99.7&99.6&\textbf{64.2}&\textbf{64.5} &\textbf{97.3}&\textbf{95.5}&\textbf{85.9}&\textbf{84.7}&\textbf{75.8}&\textbf{56.9}&\textbf{39.2}&\textbf{28.7}\\
\hline
\end{tabular}
}
\label{tab:cc_compare}
\end{table*}

\begin{table}[!h]
\small
\caption{Comparison of CSCL and other competitors on Market-1501 (single-query setting) and DukeMTMC.}
\renewcommand{\arraystretch}{1.2}

\centering
\resizebox{0.37\textwidth}{!}{
\begin{tabular}{c|cc|cc}
\hline
\multirow{2}{*}{Methods}& \multicolumn{2}{c|}{Market-1501} & \multicolumn{2}{c}{DukeMTMC}\\
\cline{2-5}
\multirow{1}{*}{} & Rank-1 & mAP &Rank-1&mAP\\
\hline
PCB (ECCV18)~\cite{GeneralPerson3}    &92.3&77.4&81.8&66.1  \\
HACNN (CVPR18)~\cite{HACNN}   &   91.2&75.7&80.5&63.8    \\
MGN (MM18)~\cite{MGN} &\textbf{95.7}&86.9&88.7&78.4\\
Trans-ReID (ICCV21)~\cite{TransReID} &95.2&\textbf{89.5}&\textbf{90.7}&82.6\\
\hline
3DSL (CVPR21)~\cite{3DSL}&95.0&87.3&88.2&76.1\\
Baseline(ResNet-50)   &    92.7  &    78.0&85.8&75.3       \\
\textbf{CSCL}   &    95.4  &    \textbf{89.5}&90.3&\textbf{83.1}       \\
\hline
\end{tabular}
}
\vspace{-0.4cm}
\label{tab:same_compare}
\end{table}

\section{Experiments}
\subsection{Datasets and Protocals}
\label{subsec:protocal}
We conduct experiments on four existing cloth-changing ReID datasets (i.e. LTCC~\cite{LTCC2020}, PRCC~\cite{PRCC}, VC-Clothes~\cite{VCClothes} and DP3D). Furthermore, three different settings are involved in our experiment: (1) \textbf{Standard Setting}: the test set includes both same-appearance and cross-appearance samples; (2) \textbf{Cloth-Changing Setting}: the test set only includes cross-appearance samples; (3) \textbf{Same-Clothes Setting}: the test set only includes same-appearance samples. For LTCC and DP3D, we provide experimental results in the standard setting and cloth-changing setting, while for PRCC and VC-Clothes, results in the same-clothes setting and cloth-changing setting are reported. We additionally validate our method on two general ReID datasets (i.e. Market-1501~\cite{Market2015} and DukeMTMC~\cite{DukeMTMC}), following their evaluation metrics. For evaluation, we adopt the mean average precision (mAP) and rank-1 accuracy to evaluate the effectiveness of ReID methods. We also utilize Geodesic Point Similarity (GPS)~\cite{DensePose2018} scores to measure the quality of the established correspondences:
\begin{equation}
    GPS_I = \frac{1}{N}\sum_{p\in I} {exp \frac{-g(v_p, \hat{v_p})^2}{2\sigma^2}}
\end{equation}
where I indicates a person image, N is the number of ground-truth correspondences, $v_p$ and $\hat{v_p}$ denote the ground-truth vertex and the estimated vertex, $g(\cdot, \cdot)$ represents geodesic distances, and $\sigma$ is a normalizing factor set to 0.255. When GPS scores exceed a certain threshold, the correspondences are considered as correct. Therefore, following the metric of BodyMap~\cite{BodyMap2022}, we report Average Precision (AP) and Average Recall (AR) based on GPS scores.
\subsection{Implementation Details}
For datasets without ground-truth dense correspondences, we fit the SMPL body model to the person images under the guidance of OpenPose~\cite{OpenPose} keypoint detections and foreground silhouettes. For each SMPL mesh vertex, there is a reprojected point on the 2D image plane, and the pixel closest to this point is utilized to establish the correspondence. If different vertices correspond to the same pixel, only the vertex closest to the camera is recorded. Based on the image resolution, we uniformly sampled 80 to 125 pseudo correspondences within the entire body region. All input images are resized to 256$\times$128. A skip-connecting UNet~\cite{UNet} architecture pretrained on the DensePose-COCO dataset~\cite{DensePose2018} is employed as embedding layers, while two distinct ResNet-50 backbone pretrained on ImageNet~\cite{ImageNet} with the last downsampling layer discarded are employed to extract global RGB features and shape features, respectively. In the training stage, the Adam optimizer\cite{Adam} was utilized for optimization. We first trained the embedding layers for 50 epochs with a learning rate of $5\times10^{-5}$, and then fixed them to train the rest of the network for 100 epochs with a linear warm-up phase. The learning rate was increased from $1\times10^{-5}$ to $1\times10^{-4}$ in the first 5 epochs. Finally, we trained the network in an end-to-end manner for 40 epochs with a fixed learning rate of $1\times10^{-5}$. The embedding dimension $D$ is set to 64. The values of $\lambda_1$, $\alpha$, $\lambda_2$, $\lambda_3$ in Eq.~\ref{eq:floss} are set to 0.3, 5.0, 1.0, 0.8, and the margin parameter for the triplet loss is set to 0.3, respectively.

\subsection{Comparison with State-of-the-arts}
As shown in Table\ref{tab:cc_compare}, we compare our proposed CSCL with seven SOTA cloth-changing methods (i.e. SE+CSED~\cite{LTCC2020}, PSAM~\cite{ShapeAppearance2021}, 3DSL~\cite{3DSL}, UCAD~\cite{Weakening2022}, MVSE~\cite{multigranular2022}, M2NET~\cite{NKUP+} and CAL~\cite{RGBOnly2022}) on LTCC, PRCC, VC-Clothes, and DP3D. To assess the feasibility of CSCL in cases without clothing change, we also choose four SOTA short-term methods (i.e. PCB~\cite{GeneralPerson3}, HACNN~\cite{HACNN}, MGN~\cite{MGN}, and Trans-ReID~\cite{TransReID}) as competitors. The comparative results on the Market-1501 and DukeMTMC are presented in Table \ref{tab:same_compare}.
   
   Based on the results in Table \ref{tab:cc_compare} and Table\ref{tab:same_compare}, we have the following key observations: (1) In the cloth-changing setting, CSCL exceeds other competitors on PRCC, VC-Clothes, and DP3D by a large margin, achieving a rank-1 improvement of 4.9\%/3.5\%/9.6\% and a mAP improvement of 6.8\%/3.5\%/10.9\%. This is attributed to the powerful shape representation capability of the continuous correspondences. However, there is still a limitation to CSCL. Due to the poor quality of person images, the generated pseudo correspondences on LTCC are not reliable enough. Despite this limitation, CSCL still achieves comparable results with the SOTA method MVSE on LTCC, indicating a certain tolerance for vertex position errors. (2) CSCL generalizes well to the general ReID datasets where appearance features dominate, achieving comparable performance with the SOTA short-term methods. This is because the distribution of global RGB features is well preserved in the fusion stage.
\vspace{-0.1cm}
\begin{table}[!t]
\normalsize
\caption{Ablation studies of different components in the CSCL framework. LNP represents linear projection, PE denotes positional embeddings, and SEN denotes the shape extraction network, respectively.}
\renewcommand{\arraystretch}{1.2}
\centering
\resizebox{0.487\textwidth}{!}{
\begin{tabular}{c|cc|c|ccc|cc|cc}
\hline
\multirow{2}{*}{Models}&\multirow{2}{*}{CSE}&\multirow{2}{*}{SEN}&\multirow{2}{*}{CMF}&\multicolumn{3}{c|}{Projection}&\multicolumn{2}{c|}{PRCC}&\multicolumn{2}{c}{DP3D}\\
\cline{5-11}
\multirow{1}{*}{}&\multirow{1}{*}{}&\multirow{1}{*}{}&\multirow{1}{*}{}&LNP+PE&CP&LCP& Rank-1 & mAP &Rank-1&mAP\\
\hline
1(Baseline)&\usym{2717}&-&-&-&-&-&32.8&37.1&19.2&13.0\\
2&\usym{2713}&\usym{2717}&\usym{2717}&-&-&-&34.2&38.8&21.9&14.2\\
3&\usym{2713}&\usym{2713}&\usym{2717}&-&-&-&52.9&55.4&30.7&23.5\\
4&\usym{2713}&\usym{2713}&\usym{2713}&\usym{2713}&\usym{2717}&\usym{2717}&62.5&63.7&37.7&27.1\\
5&\usym{2713}&\usym{2713}&\usym{2713}&\usym{2717}&\usym{2713}&\usym{2717}&62.8&63.7&37.7&27.3\\
6&\usym{2713}&\usym{2713}&\usym{2713}&\usym{2717}&\usym{2717}&\usym{2713}&\textbf{64.2}&\textbf{64.5}&\textbf{39.2}&\textbf{28.7}\\

\hline
\end{tabular}
}
\vspace{-0.4cm}
\label{tab:Ablstudy}
\end{table}
\subsection{Ablation Studies}
In this section, we carry out comprehensive experiments on PRCC and DP3D to validate: (1) the effectiveness of continuous surface embeddings, the cross-modality fusion module, and latent convolutional projection, which are abbreviated as CSE, CMF, and LCP respectively; (2) the influence of consistency loss on correspondence learning;  (3) the impact of using different features for inference.

\begin{table}[!t]
\small
\caption{Average Precision (AP) and Recall (AR) calculated at GPS thresholds ranging from 0.5 to 0.95 on multiple datasets.}
\renewcommand{\arraystretch}{1.1}
\centering
\resizebox{0.44\textwidth}{!}{
\begin{tabular}{c|cccccc}
\hline
Datasets&$AP_{50}$&$AP_{75}$&$AP_{95}$&$AR_{50}$&$AR_{75}$&$AR_{95}$\\
\hline
Market-1501&67.6&58.5&50.8&70.0&60.8&52.4\\
DukeMTMC&63.1&52.3&45.6&63.5&53.1&46.0\\
LTCC&59.2&49.8&39.5&60.3&51.7&39.8\\
PRCC&66.4&57.4&49.7&67.6&59.3&50.9\\
VC-Clothes&73.0&64.9&59.2&74.1&67.1&58.8\\
DP3D (w/o. $\mathcal{L}_{cst}$)&84.0&74.6&65.8&84.9&76.0&66.2\\
DP3D&\textbf{87.5}&\textbf{79.6}&\textbf{70.3}&\textbf{90.3}&\textbf{81.2}&\textbf{70.5}\\
\hline
\end{tabular}
}
\vspace{-0.2cm}
\label{tab:cse_compare}
\end{table}
\begin{table}[!t]
\small
\caption{Ablation studies of deploying different features for inference. CLS denotes the two learnable class tokens.}
\renewcommand{\arraystretch}{1.1}
\centering
\resizebox{0.35\textwidth}{!}{
\begin{tabular}{c|cc|cc}
\hline
\multirow{2}{*}{Features}&\multicolumn{2}{c|}{PRCC}&\multicolumn{2}{c}{DP3D}\\
\cline{2-5}
\multirow{1}{*}{} &Rank-1 & mAP &Rank-1&mAP\\
\hline
RGB&55.9&57.7&36.8&26.3\\
CLS&61.1&61.8&37.9&27.4\\
Shape&42.5&45.9&23.9&16.8\\
RGB + Shape&61.4&62.6&37.4&27.2\\
CLS + Shape&63.5&64.2&38.7&28.4\\
RGB +  CLS + Shape&\textbf{64.2}&\textbf{64.5}&\textbf{39.2}&\textbf{28.7}\\
\hline
\end{tabular}
}
\vspace{-0.3cm}
\label{tab:feature_compare}
\end{table}
\textbf{Effectiveness of CSE, CMF, and LCP.} 
From Table~\ref{tab:Ablstudy}, we observe that introducing continuous surface embeddings to the model with a proper shape extraction network (Baseline$\rightarrow$Model3) remarkably improves the performance of the baseline model, with a rank-1/mAP improvement of 20.1\%/18.3\% on PRCC, and a rank-1/mAP improvement of 11.5\%/10.5\% on DP3D. This demonstrates that establishing pixel-wise and continuous correspondences complement rich and essential identity-related shape features for global RGB features. However, there is no significant improvement when directly downsampling the learned correspondences without a shape extraction network, and we will further analyze this issue in Section~\ref{subsec:analy}. Moreover, the cross-modality fusion module also brings significant improvement, which indicates that features of the two modalities become more compatible via cross-modality fusion. Furthermore, by comparing different feature projection methods for generating Q/K/V, we observe that LCP shows a certain degree of improvement over linear projection and convolutional projection. This is attributed to the inclusion of latent embeddings, which greatly facilitates the sharing among tokens. 

Additionally, we evaluate the quality of established correspondences on different ReID datasets in Table~\ref{tab:cse_compare}. By combining the results from Table~\ref{tab:cc_compare} and Table~\ref{tab:cse_compare}, we can clearly observe a robust positive correlation between the quality of correspondences and the magnitude of performance improvement.

\textbf{Influence of consistency loss.} As shown in Table~\ref{tab:cse_compare}, the removal of consistency loss $\mathcal{L}_{cst}$ from the correspondence learning process leads to a 5\% decrease in vertex classification accuracy on DP3D, which indicates that performing consistency learning is beneficial for establishing reliable correspondences. From Table ~\ref{tab:cc_compare}, we also observe that removing $\mathcal{L}_{cst}$ results in a decline in the overall performance of ReID, verifying the importance of consistency learning for CSE.

\textbf{Impact of using different features for inference.} 
 During inference, we select the model corresponding to Model 6 in Table~\ref{tab:Ablstudy} to verify the effectiveness of different features. As shown in Table~\ref{tab:feature_compare}, while relying solely on shape features is not reliable enough, the shape features can enhance the performance of other features. Concatenating global RGB features, shape features, and two class tokens results in the best performance at inference time.

\subsection{Further Analysis}
\label{subsec:analy}
\begin{figure}[!t]
    \centering
    \includegraphics[width=0.48\textwidth]{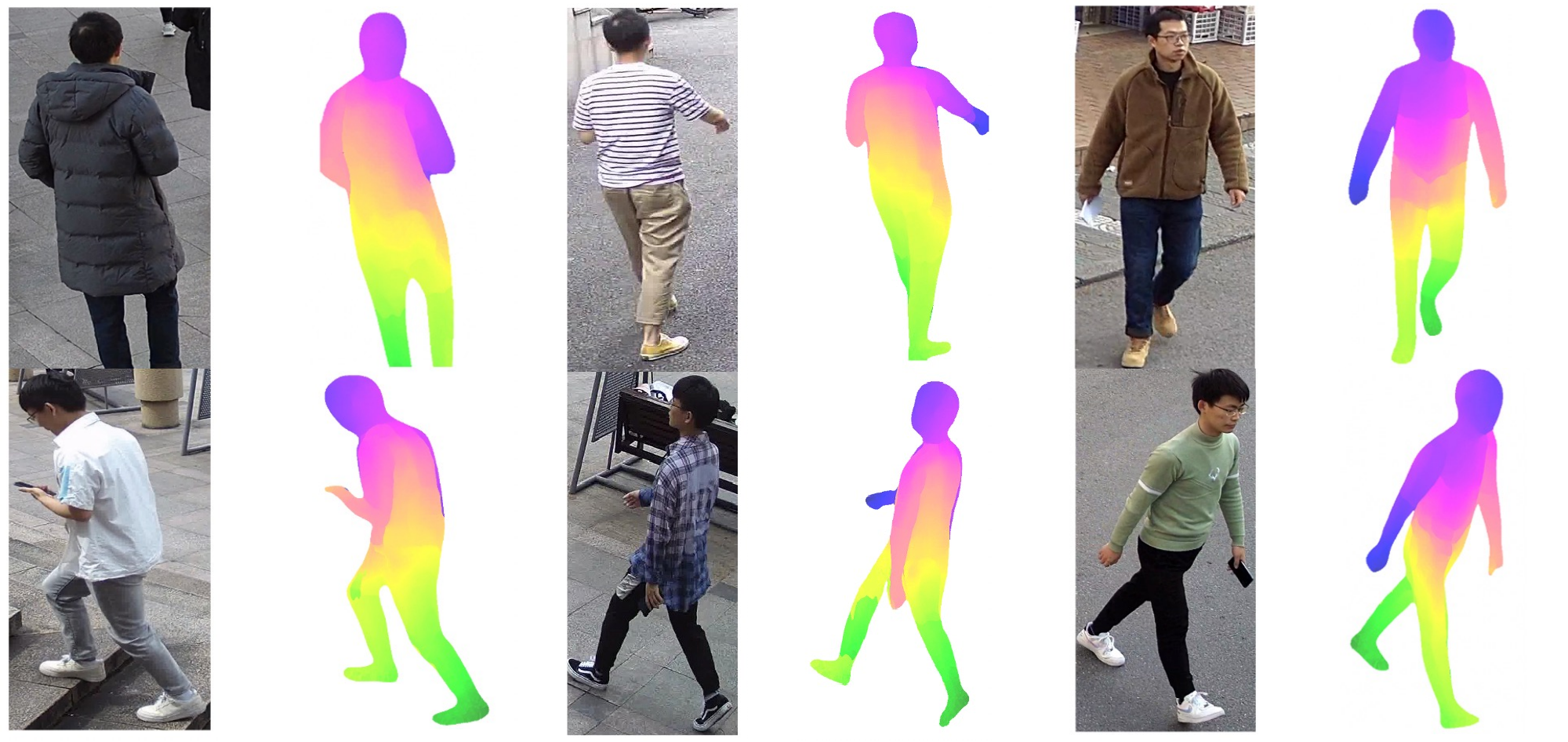}
    \caption{PCA visualization results of the learned continuous surface embeddings. The person images in each row are cross-appearance images of the same person in DP3D. We reduce the channel dimension of the learned continuous surface embeddings from 64 to 3 for visualization. }
    \label{fig:pca}
\vspace{-0.7cm}
\end{figure}

\textbf{Visualization of Continuous Surface Embeddings.}
We employ PCA to reduce the dimension of continuous surface embeddings from $H\times W\times D$ to $H\times W \times 3$, where $H$ and $W$ denote the height and width of person images, $D$ represents the embedding dimension. Visualization results on DP3D are presented in Figure~\ref{fig:pca}. Since the color differences reflect the feature distances in the embedding space, we can clearly observe that the established 2D-3D correspondences between images pixels and the entire body surface are relatively smooth. Different from discretized UV mappings such as the DensePose, the smooth and continuous 2D-3D correspondences can provide richer and more reliable global knowledge of human shape for cloth-changing ReID.

\textbf{Identity modeling for shape features.}  Multi-modal auxiliary information itself is not sufficiently discriminative for the ReID task, making it necessary to conduct identity modeling. However, some existing CC-ReID methods, such as 3DSL, directly regulate multi-modal auxiliary features via ReID losses, which disrupts the distribution of shape space. As shown in Table~\ref{tab:Ablstudy}, directly using downsampling operations without a proper shape extraction network (Model3$\rightarrow$Model2) leads to significant performance degradation. We believe that multi-modal auxiliary features should first be mapped to an intermediary feature space before identity modeling to alleviate the incompatibility between feature spaces of different tasks, which is beneficial for the fusion of shape and global RGB features.

\textbf{Future works.} Current 3D shape-based ReID methods suffer from a huge domain gap between the RGB image space and the 3D shape space. Our work essentially targets at bridging the gap between these two spaces. Therefore, future works can consider transforming the surface mebddings into different forms of 3D shape features and assess their potential benefits for CC-ReID.

\section{Conclusion}
We have proposed a new shape embedding paradigm that establishes pixel-wise and continuous surface correspondences to mine fine-grained shape features for cloth-changing ReID. Moreover, an optimized cross-modality fusion module is designed to adaptively integrate shape features with global RGB features. To facilitate the research, we have constructed 3D Dense Persons (DP3D), which is the first cloth-changing ReID dataset with densely annotated 2D-3D correspondences and corresponding 3D meshes. Experiments on both cloth-changing and cloth-consistent ReID benchmarks demonstrate the robustness and superiority of our method.
\begin{acks}
This work was supported in part by the Research Project of ZJU-League Research \& Development Center, Zhejiang Lab under Grant 2019KD0AB01.
\end{acks}
\bibliographystyle{ACM-Reference-Format}
\balance
\bibliography{CSCL}


\begin{thebibliography}{56}


\ifx \showCODEN    \undefined \def \showCODEN     #1{\unskip}     \fi
\ifx \showDOI      \undefined \def \showDOI       #1{#1}\fi
\ifx \showISBNx    \undefined \def \showISBNx     #1{\unskip}     \fi
\ifx \showISBNxiii \undefined \def \showISBNxiii  #1{\unskip}     \fi
\ifx \showISSN     \undefined \def \showISSN      #1{\unskip}     \fi
\ifx \showLCCN     \undefined \def \showLCCN      #1{\unskip}     \fi
\ifx \shownote     \undefined \def \shownote      #1{#1}          \fi
\ifx \showarticletitle \undefined \def \showarticletitle #1{#1}   \fi
\ifx \showURL      \undefined \def \showURL       {\relax}        \fi
\providecommand\bibfield[2]{#2}
\providecommand\bibinfo[2]{#2}
\providecommand\natexlab[1]{#1}
\providecommand\showeprint[2][]{arXiv:#2}

\bibitem[Alp~Güler et~al\mbox{.}(2018)]%
        {DensePose2018}
\bibfield{author}{\bibinfo{person}{Rıza Alp~Güler}, \bibinfo{person}{Natalia
  Neverova}, {and} \bibinfo{person}{Kokkinos Iasonas}.}
  \bibinfo{year}{2018}\natexlab{}.
\newblock \showarticletitle{DensePose: Dense Human Pose Estimation in the
  Wild}. In \bibinfo{booktitle}{\emph{Proceedings of the IEEE/CVF Conference on
  Computer Vision and Pattern Recognition}}. \bibinfo{pages}{7297--7306}.
\newblock


\bibitem[Cao et~al\mbox{.}(2021)]%
        {OpenPose}
\bibfield{author}{\bibinfo{person}{Zhe Cao}, \bibinfo{person}{Tomas Simon},
  \bibinfo{person}{Shih-En Wei}, {and} \bibinfo{person}{Yaser Sheikh}.}
  \bibinfo{year}{2021}\natexlab{}.
\newblock \showarticletitle{Realtime Multi-Person 2D Pose Estimation Using Part
  Affinity Fields}. In \bibinfo{booktitle}{\emph{Proceedings of the IEEE/CVF
  Conference on Computer Vision and Pattern Recognition}}.
  \bibinfo{pages}{7291--7299}.
\newblock


\bibitem[Chan et~al\mbox{.}(2023)]%
        {Distengled2023}
\bibfield{author}{\bibinfo{person}{Patrich P.~K. Chan}, \bibinfo{person}{Haorui
  Song}, \bibinfo{person}{Peng Peng}, \bibinfo{person}{Keke Chen}, {and}
  \bibinfo{person}{Daniel S.Yeung}.} \bibinfo{year}{2023}\natexlab{}.
\newblock \showarticletitle{Learning Disentangled Features for Person
  Re-Identification under Clothes Changing}.
\newblock \bibinfo{journal}{\emph{ACM Transactions on Multimedia Computing,
  Communications, and Applications.}}  \bibinfo{volume}{DOI:10.1145/3584359}
  (\bibinfo{year}{2023}).
\newblock


\bibitem[Chen et~al\mbox{.}(2021)]%
        {3DSL}
\bibfield{author}{\bibinfo{person}{Jiaxing Chen}, \bibinfo{person}{Xinyang
  Jiang}, \bibinfo{person}{Fudong Wang}, \bibinfo{person}{Jun Zhang},
  \bibinfo{person}{Feng Zheng}, \bibinfo{person}{Xing Sun}, {and}
  \bibinfo{person}{Wei-Shi Zheng}.} \bibinfo{year}{2021}\natexlab{}.
\newblock \showarticletitle{Learning 3D Shape Feature for Texture-Insensitive
  Person Re-Identification}. In \bibinfo{booktitle}{\emph{Proceedings of the
  IEEE/CVF Conference on Computer Vision and Pattern Recognition}}.
  \bibinfo{pages}{8146--8155}.
\newblock


\bibitem[Deng et~al\mbox{.}(2009)]%
        {ImageNet}
\bibfield{author}{\bibinfo{person}{Jia Deng}, \bibinfo{person}{Wei Dong},
  \bibinfo{person}{Richard Socher}, \bibinfo{person}{Li-Jia Li},
  \bibinfo{person}{Kai Li}, {and} \bibinfo{person}{Fei-Fei Li}.}
  \bibinfo{year}{2009}\natexlab{}.
\newblock \showarticletitle{Imagenet: A Large-Scale Hierarchical Image
  Database}. In \bibinfo{booktitle}{\emph{Proceedings of the IEEE/CVF
  Conference on Computer Vision and Pattern Recognition}}.
\newblock


\bibitem[Farenzena et~al\mbox{.}(2010)]%
        {GeneralPerson4}
\bibfield{author}{\bibinfo{person}{Michela Farenzena}, \bibinfo{person}{Loris
  Bazzani}, \bibinfo{person}{Alessandro Perina}, \bibinfo{person}{Vittorio
  Murino}, {and} \bibinfo{person}{Marco Cristani}.}
  \bibinfo{year}{2010}\natexlab{}.
\newblock \showarticletitle{Person Re-Identification by Symmetry-Driven
  Accumulation of Local Features}. In \bibinfo{booktitle}{\emph{Proceedings of
  the IEEE/CVF Conference on Computer Vision and Pattern Recognition}}.
  \bibinfo{pages}{2360--2367}.
\newblock


\bibitem[Gao et~al\mbox{.}(2022)]%
        {multigranular2022}
\bibfield{author}{\bibinfo{person}{Zan Gao}, \bibinfo{person}{Hongwei Wei},
  \bibinfo{person}{Weili Guan}, \bibinfo{person}{Weizhi Nei},
  \bibinfo{person}{Meng Liu}, {and} \bibinfo{person}{Meng Wang}.}
  \bibinfo{year}{2022}\natexlab{}.
\newblock \showarticletitle{Multigranular Visual-Semantic Embedding for
  Cloth-Changing Person Re-identification}. In
  \bibinfo{booktitle}{\emph{Proceedings of the 30th ACM International
  Conference on Multimedia}}. \bibinfo{pages}{3703--3711}.
\newblock


\bibitem[Gong et~al\mbox{.}(2019)]%
        {Graphonomy}
\bibfield{author}{\bibinfo{person}{Ke Gong}, \bibinfo{person}{Yiming Gao},
  \bibinfo{person}{Xiaodan Liang}, \bibinfo{person}{Xiaohui Shen},
  \bibinfo{person}{Meng Wang}, {and} \bibinfo{person}{Liang Lin}.}
  \bibinfo{year}{2019}\natexlab{}.
\newblock \showarticletitle{Graphonomy: Universal Human Parsing via Graph
  Transfer Learning}. In \bibinfo{booktitle}{\emph{Proceedings of the IEEE/CVF
  International Conference on Computer Vision}}. \bibinfo{pages}{7450--7459}.
\newblock


\bibitem[Gu et~al\mbox{.}(2022)]%
        {RGBOnly2022}
\bibfield{author}{\bibinfo{person}{Xinqian Gu}, \bibinfo{person}{Hong Chang},
  \bibinfo{person}{Bingpeng Ma}, \bibinfo{person}{Shutao Bai},
  \bibinfo{person}{Shiguang Shan}, {and} \bibinfo{person}{Xilin Chen}.}
  \bibinfo{year}{2022}\natexlab{}.
\newblock \showarticletitle{Clothes-Changing Person Re-identification with RGB
  Modality Only}. In \bibinfo{booktitle}{\emph{Proceedings of the IEEE/CVF
  Conference on Computer Vision and Pattern Recognition}}.
  \bibinfo{pages}{1060--1069}.
\newblock


\bibitem[Haugaard and Buch(2022)]%
        {SurfEmb}
\bibfield{author}{\bibinfo{person}{Rasmus~Laurvig Haugaard} {and}
  \bibinfo{person}{Anders~Glent Buch}.} \bibinfo{year}{2022}\natexlab{}.
\newblock \showarticletitle{SurfEmb: Dense and Continuous Correspondence
  Distributions for Object Pose Estimation With Learnt Surface Embeddings}. In
  \bibinfo{booktitle}{\emph{Proceedings of the IEEE/CVF Conference on Computer
  Vision and Pattern Recognition}}. \bibinfo{pages}{6749--6758}.
\newblock


\bibitem[He et~al\mbox{.}(2017)]%
        {MaskRCNN2017}
\bibfield{author}{\bibinfo{person}{Kaiming He}, \bibinfo{person}{Georgia
  Gkioxari}, \bibinfo{person}{Piotr Dollar}, {and} \bibinfo{person}{Ross
  Girshick}.} \bibinfo{year}{2017}\natexlab{}.
\newblock \showarticletitle{Mask R-CNN}. In
  \bibinfo{booktitle}{\emph{Proceedings of the IEEE/CVF International
  Conference on Computer Vision}}. \bibinfo{pages}{2961--2969}.
\newblock


\bibitem[He et~al\mbox{.}(2016)]%
        {HeRes}
\bibfield{author}{\bibinfo{person}{Kaiming He}, \bibinfo{person}{Xiangyu
  Zhang}, \bibinfo{person}{Shaoqing Ren}, {and} \bibinfo{person}{Jian Sun}.}
  \bibinfo{year}{2016}\natexlab{}.
\newblock \showarticletitle{Deep Residual Learning for Image Recognition}. In
  \bibinfo{booktitle}{\emph{Proceedings of the IEEE/CVF Conference on Computer
  Vision and Pattern Recognition}}. \bibinfo{pages}{770--778}.
\newblock


\bibitem[He et~al\mbox{.}(2021)]%
        {TransReID}
\bibfield{author}{\bibinfo{person}{Shuting He}, \bibinfo{person}{Hao Luo},
  \bibinfo{person}{Pichao Wang}, \bibinfo{person}{Fang Wang},
  \bibinfo{person}{Hao Li}, {and} \bibinfo{person}{Wei Jiang}.}
  \bibinfo{year}{2021}\natexlab{}.
\newblock \showarticletitle{TransReID: Transformer-Based Object
  Re-Identification}. In \bibinfo{booktitle}{\emph{Proceedings of the IEEE/CVF
  International Conference on Computer Vision}}. \bibinfo{pages}{15013--15022}.
\newblock


\bibitem[Hermans et~al\mbox{.}(2017)]%
        {triplet}
\bibfield{author}{\bibinfo{person}{Alexander Hermans}, \bibinfo{person}{Lucas
  Beyer}, {and} \bibinfo{person}{Bastian Leibe}.}
  \bibinfo{year}{2017}\natexlab{}.
\newblock \showarticletitle{In Defense of the Triplet Loss for Person
  Re-Identification}.
\newblock   \bibinfo{volume}{arXiv preprint arXiv:1703.07737}
  (\bibinfo{year}{2017}).
\newblock


\bibitem[Hong et~al\mbox{.}(2021)]%
        {ShapeAppearance2021}
\bibfield{author}{\bibinfo{person}{Peixian Hong}, \bibinfo{person}{Tao Wu},
  \bibinfo{person}{Ancong Wu}, \bibinfo{person}{Xintong Han}, {and}
  \bibinfo{person}{Wei-Shi Zheng}.} \bibinfo{year}{2021}\natexlab{}.
\newblock \showarticletitle{Fine-Grained Shape-Appearance Mutual Learning for
  Cloth-Changing Person Re-Identification}. In
  \bibinfo{booktitle}{\emph{Proceedings of the IEEE/CVF Conference on Computer
  Vision and Pattern Recognition}}. \bibinfo{pages}{10513--10522}.
\newblock


\bibitem[Huang et~al\mbox{.}(2019a)]%
        {Celeb2019}
\bibfield{author}{\bibinfo{person}{Yan Huang}, \bibinfo{person}{Jingsong Xu},
  \bibinfo{person}{Qiang Wu}, \bibinfo{person}{Yi Zhong}, \bibinfo{person}{Peng
  Zhang}, {and} \bibinfo{person}{zhaoxiang Zhang}.}
  \bibinfo{year}{2019}\natexlab{a}.
\newblock \showarticletitle{Beyond Scalar Neuron: Adopting Vector-Neuron
  Capsules for Long-Term Person Re-Identification}.
\newblock \bibinfo{journal}{\emph{IEEE Transactions on Circuits and Systems for
  Video Technology (TCSVT)}} \bibinfo{volume}{30}, \bibinfo{number}{10}
  (\bibinfo{year}{2019}), \bibinfo{pages}{3459--3471}.
\newblock


\bibitem[Huang et~al\mbox{.}(2019b)]%
        {General6}
\bibfield{author}{\bibinfo{person}{Yukun Huang}, \bibinfo{person}{Zheng-Jun
  Zha}, \bibinfo{person}{Xueyang Fu}, {and} \bibinfo{person}{Wei Zhang}.}
  \bibinfo{year}{2019}\natexlab{b}.
\newblock \showarticletitle{Illumination-Invariant Person Re-Identification}.
  In \bibinfo{booktitle}{\emph{Proceedings of the 27th ACM International
  Conference on Multimedia}}. \bibinfo{pages}{365--373}.
\newblock


\bibitem[Ianina et~al\mbox{.}(2022)]%
        {BodyMap2022}
\bibfield{author}{\bibinfo{person}{Anastasia Ianina}, \bibinfo{person}{Nikolaos
  Sarafianos}, \bibinfo{person}{Yuanlu Xu}, \bibinfo{person}{Ignacio Rocco},
  {and} \bibinfo{person}{Tony Tung}.} \bibinfo{year}{2022}\natexlab{}.
\newblock \showarticletitle{BodyMap: Learning Full-Body Dense Correspondence
  Map}. In \bibinfo{booktitle}{\emph{Proceedings of the IEEE/CVF Conference on
  Computer Vision and Pattern Recognition}}. \bibinfo{pages}{13286--13295}.
\newblock


\bibitem[Jia et~al\mbox{.}(2022)]%
        {Patching2022}
\bibfield{author}{\bibinfo{person}{Xuemei Jia}, \bibinfo{person}{Xian Zhong},
  \bibinfo{person}{Mang Ye}, \bibinfo{person}{Wenxuan Liu},
  \bibinfo{person}{Xenxin Huang}, {and} \bibinfo{person}{Shilei Zhao}.}
  \bibinfo{year}{2022}\natexlab{}.
\newblock \showarticletitle{Patching Your Clothes: Semantic-Aware Learning for
  Cloth-Changed Person Re-Identification}. In
  \bibinfo{booktitle}{\emph{International Conference on Multimedia Modeling
  (MMM)}}. \bibinfo{pages}{121--133}.
\newblock


\bibitem[Jiao et~al\mbox{.}(2022)]%
        {VAF}
\bibfield{author}{\bibinfo{person}{Bingliang Jiao}, \bibinfo{person}{Lingqiao
  Liu}, \bibinfo{person}{Liying Gao}, \bibinfo{person}{Guosheng Lin},
  \bibinfo{person}{Ruiqi Wu}, \bibinfo{person}{Shizhou Zhang},
  \bibinfo{person}{Peng Wang}, {and} \bibinfo{person}{Yanning Zhang}.}
  \bibinfo{year}{2022}\natexlab{}.
\newblock \showarticletitle{Generalizable Person Re-Identification via
  Viewpoint Alignment and Fusion}.
\newblock   \bibinfo{volume}{arXiv preprint arXiv:2212.02398}
  (\bibinfo{year}{2022}).
\newblock


\bibitem[Jin et~al\mbox{.}(2022)]%
        {GaitCC}
\bibfield{author}{\bibinfo{person}{Xin Jin}, \bibinfo{person}{Tianyu He},
  \bibinfo{person}{Kecheng Zheng}, \bibinfo{person}{Zhiheng Yin},
  \bibinfo{person}{Xu Shen}, \bibinfo{person}{Zhen Huang},
  \bibinfo{person}{Ruoyu Feng}, \bibinfo{person}{Jianqiang Huang},
  \bibinfo{person}{Zhibo Chen}, {and} \bibinfo{person}{Xian-Sheng Hua}.}
  \bibinfo{year}{2022}\natexlab{}.
\newblock \showarticletitle{Cloth-Changing Person Re-identification from A
  Single Image with Gait Prediction and Regularization}. In
  \bibinfo{booktitle}{\emph{Proceedings of the IEEE/CVF Conference on Computer
  Vision and Pattern Recognition}}. \bibinfo{pages}{14278--14287}.
\newblock


\bibitem[Li et~al\mbox{.}(2022)]%
        {AlignTransformer}
\bibfield{author}{\bibinfo{person}{Hui Li}, \bibinfo{person}{Yinglin Zheng},
  \bibinfo{person}{Zhaodong Tan}, {and} \bibinfo{person}{Wenjin Deng}.}
  \bibinfo{year}{2022}\natexlab{}.
\newblock \showarticletitle{Improving Person Re-identification with
  Semantically Aligned Appearance Transformer}. In
  \bibinfo{booktitle}{\emph{International Joint Conference on Neural
  Networks}}.
\newblock


\bibitem[Li et~al\mbox{.}(2020)]%
        {General7}
\bibfield{author}{\bibinfo{person}{Shuzhao Li}, \bibinfo{person}{Huimin Yu},
  {and} \bibinfo{person}{Roland Hu}.} \bibinfo{year}{2020}\natexlab{}.
\newblock \showarticletitle{Attributes-aided Part Detection and Refinement for
  Person Re-Identification}.
\newblock \bibinfo{journal}{\emph{Pattern Recognition}}  \bibinfo{volume}{97}
  (\bibinfo{year}{2020}), \bibinfo{pages}{4326--4335}.
\newblock


\bibitem[Li et~al\mbox{.}(2014)]%
        {CUKE03}
\bibfield{author}{\bibinfo{person}{Wei Li}, \bibinfo{person}{Rui Zhao},
  \bibinfo{person}{Tong Xiao}, {and} \bibinfo{person}{Xiaogang Wang}.}
  \bibinfo{year}{2014}\natexlab{}.
\newblock \showarticletitle{DeepReID: Deep Filter Pairing Neural Network for
  Person Re-Identification}. \bibinfo{pages}{152--159}.
\newblock


\bibitem[Li et~al\mbox{.}(2018)]%
        {HACNN}
\bibfield{author}{\bibinfo{person}{Wei Li}, \bibinfo{person}{Xiatian Zhu},
  {and} \bibinfo{person}{Shangang Gong}.} \bibinfo{year}{2018}\natexlab{}.
\newblock \showarticletitle{Harmonious Attention Network for Person
  Re-Identification}. In \bibinfo{booktitle}{\emph{Proceedings of the IEEE/CVF
  Conference on Computer Vision and Pattern Recognition}}.
  \bibinfo{pages}{2285--2294}.
\newblock


\bibitem[Liu et~al\mbox{.}(2022)]%
        {NKUP+}
\bibfield{author}{\bibinfo{person}{Mengmeng Liu}, \bibinfo{person}{Zhi Ma},
  \bibinfo{person}{Tao Li}, \bibinfo{person}{Yanfeng Jiang}, {and}
  \bibinfo{person}{Kai Wang}.} \bibinfo{year}{2022}\natexlab{}.
\newblock \showarticletitle{Long-Term Person Re-identification with Dramatic
  Appearance Change: Algorithm and Benchmark}. In
  \bibinfo{booktitle}{\emph{Proceedings of the 30th ACM International
  Conference on Multimedia}}. \bibinfo{pages}{6406--6415}.
\newblock


\bibitem[Loper et~al\mbox{.}(2015)]%
        {smpl2015}
\bibfield{author}{\bibinfo{person}{Matthew Loper}, \bibinfo{person}{Naureen
  Mahmood}, \bibinfo{person}{Javier Romero}, \bibinfo{person}{Gerard
  Pons-Moll}, {and} \bibinfo{person}{Michael J~Black}.}
  \bibinfo{year}{2015}\natexlab{}.
\newblock \showarticletitle{SMPL: A Skinned Multi-Person Linear Model}.
\newblock \bibinfo{journal}{\emph{ACM transactions on graphics (TOG)}}
  \bibinfo{volume}{34}, \bibinfo{number}{6} (\bibinfo{year}{2015}).
\newblock


\bibitem[Ming et~al\mbox{.}(2022a)]%
        {surv2022}
\bibfield{author}{\bibinfo{person}{Zhangqiang Ming}, \bibinfo{person}{Min Zhu},
  \bibinfo{person}{Xiangkun Wang}, \bibinfo{person}{Jiaming Zhu},
  \bibinfo{person}{Cheng Junlong}, \bibinfo{person}{Chengrui Gao},
  \bibinfo{person}{Yong Yang}, {and} \bibinfo{person}{Xiaoyong Wei}.}
  \bibinfo{year}{2022}\natexlab{a}.
\newblock \showarticletitle{Deep Learning-Based Person Re-Identification
  Methods: A Survey and Outlook of Recent Work}.
\newblock \bibinfo{journal}{\emph{Image and Vision Computing}}
  \bibinfo{volume}{119:104394} (\bibinfo{year}{2022}).
\newblock


\bibitem[Ming et~al\mbox{.}(2022b)]%
        {IRANet2022}
\bibfield{author}{\bibinfo{person}{Zhangqiang Ming}, \bibinfo{person}{Min Zhu},
  \bibinfo{person}{Xiangkun Wang}, \bibinfo{person}{Jiaming Zhu},
  \bibinfo{person}{Cheng Junlong}, \bibinfo{person}{Chengrui Gao},
  \bibinfo{person}{Yong Yang}, {and} \bibinfo{person}{Xiaoyong Wei}.}
  \bibinfo{year}{2022}\natexlab{b}.
\newblock \showarticletitle{IRANet: Identity-Relevance Aware Representation for
  Cloth-Changing Person Re-Identification}.
\newblock \bibinfo{journal}{\emph{Image and Vision Computing}}
  \bibinfo{volume}{117:104335} (\bibinfo{year}{2022}).
\newblock


\bibitem[Neverova et~al\mbox{.}(2020)]%
        {Continuous2020}
\bibfield{author}{\bibinfo{person}{Natalia Neverova}, \bibinfo{person}{David
  Novotny}, \bibinfo{person}{Marc Szafraniec}, \bibinfo{person}{Vasil
  Khalidov}, \bibinfo{person}{Patrick Labatut}, {and} \bibinfo{person}{Andrea
  Vedaldi}.} \bibinfo{year}{2020}\natexlab{}.
\newblock \showarticletitle{Continuous Surface Embeddings}. In
  \bibinfo{booktitle}{\emph{Proceedings of Neural Information Processing
  Systems (NeurlPS)}}. \bibinfo{pages}{17258--17270}.
\newblock


\bibitem[P~Kingma and Ba(2014)]%
        {Adam}
\bibfield{author}{\bibinfo{person}{Diederik P~Kingma} {and}
  \bibinfo{person}{Jimmy Ba}.} \bibinfo{year}{2014}\natexlab{}.
\newblock \showarticletitle{Adam: A Method for Stochastic Optimization}.
\newblock   \bibinfo{volume}{arXiv preprint arXiv:1402.6980}
  (\bibinfo{year}{2014}).
\newblock


\bibitem[Pavlakos et~al\mbox{.}(2019)]%
        {Smplify-X}
\bibfield{author}{\bibinfo{person}{Georgios Pavlakos},
  \bibinfo{person}{Vasileios Choutas}, \bibinfo{person}{Nima Ghorbani},
  \bibinfo{person}{Timo Bolkart}, \bibinfo{person}{Ahmed A.~A.~Osman},
  \bibinfo{person}{Dimitrios Tzionas}, {and} \bibinfo{person}{Michael
  J.~Black}.} \bibinfo{year}{2019}\natexlab{}.
\newblock \showarticletitle{Expressive Body Capture: 3D Hands, Face, and Body
  from a Single Image}. In \bibinfo{booktitle}{\emph{Proceedings of the
  IEEE/CVF Conference on Computer Vision and Pattern Recognition}}.
  \bibinfo{pages}{10975--10985}.
\newblock


\bibitem[Qian et~al\mbox{.}(2020)]%
        {LTCC2020}
\bibfield{author}{\bibinfo{person}{Xuelin Qian}, \bibinfo{person}{Wenxuan
  Wang}, \bibinfo{person}{Li Zhang}, \bibinfo{person}{Fangrui Zhu},
  \bibinfo{person}{Yanwei Fu}, \bibinfo{person}{Tao Xiang},
  \bibinfo{person}{Yu-Gang Jiang}, {and} \bibinfo{person}{Xiangyang Xue}.}
  \bibinfo{year}{2020}\natexlab{}.
\newblock \showarticletitle{Long-Term Cloth-Changing Person Re-Identification}.
  In \bibinfo{booktitle}{\emph{Proceedings of the Asian Conference on Computer
  Vision (ACCV)}}. \bibinfo{pages}{71--88}.
\newblock


\bibitem[Ristani et~al\mbox{.}(2016)]%
        {DukeMTMC}
\bibfield{author}{\bibinfo{person}{Ergys Ristani}, \bibinfo{person}{Francesco
  Solera}, \bibinfo{person}{Roger Zou}, \bibinfo{person}{Rita Cucchiara}, {and}
  \bibinfo{person}{Carlo Tomasi}.} \bibinfo{year}{2016}\natexlab{}.
\newblock \showarticletitle{Performance Measures and a Data Set for
  Multi-target, Multi-camera Tracking}. In
  \bibinfo{booktitle}{\emph{Proceedings of the European Conference on Computer
  Vision}}. \bibinfo{publisher}{Springer}, \bibinfo{pages}{17--35}.
\newblock


\bibitem[Ronneberger et~al\mbox{.}(2015)]%
        {UNet}
\bibfield{author}{\bibinfo{person}{Olaf Ronneberger}, \bibinfo{person}{Philipp
  Fischer}, {and} \bibinfo{person}{Thomas Brox}.}
  \bibinfo{year}{2015}\natexlab{}.
\newblock \showarticletitle{U-Net: Convolutional Networks for Biomedical Image
  Segmentation}. In \bibinfo{booktitle}{\emph{Medical Image Computing and
  Computer-Assisted Intervention (MICCAI)}}. \bibinfo{pages}{234--241}.
\newblock


\bibitem[Shu et~al\mbox{.}(2021)]%
        {SemanticGuided2021}
\bibfield{author}{\bibinfo{person}{Xiujun Shu}, \bibinfo{person}{Ge Li},
  \bibinfo{person}{Xiao Wang}, \bibinfo{person}{Weijian Ruan}, {and}
  \bibinfo{person}{Qi Tian}.} \bibinfo{year}{2021}\natexlab{}.
\newblock \showarticletitle{Semantic-Guided Pixel Sampling for Cloth-Changing
  Person Re-Identification}.
\newblock \bibinfo{journal}{\emph{IEEE Signal Process. Lett.}}
  \bibinfo{volume}{28} (\bibinfo{year}{2021}), \bibinfo{pages}{1365--1369}.
\newblock


\bibitem[Sun et~al\mbox{.}(2018)]%
        {GeneralPerson3}
\bibfield{author}{\bibinfo{person}{Yifan Sun}, \bibinfo{person}{Liang Zheng},
  \bibinfo{person}{Yi Yang}, \bibinfo{person}{Qi Tian}, {and}
  \bibinfo{person}{Shengjin Wang}.} \bibinfo{year}{2018}\natexlab{}.
\newblock \showarticletitle{Beyond Part Models: Person Retrieval with Refined
  Part Pooling (and A Strong Convolutional Baseline)}. In
  \bibinfo{booktitle}{\emph{Proceedings of the European Conference on Computer
  Vision}}. \bibinfo{publisher}{Springer}, \bibinfo{pages}{480--496}.
\newblock


\bibitem[Tan et~al\mbox{.}(2021)]%
        {HumanGPS2021}
\bibfield{author}{\bibinfo{person}{Feitong Tan}, \bibinfo{person}{Danhang
  Tang}, \bibinfo{person}{Mingsong Dou}, \bibinfo{person}{Kaiwen Guo},
  \bibinfo{person}{Rohit Pandey}, \bibinfo{person}{Cem Keskin},
  \bibinfo{person}{Ruofei Du}, \bibinfo{person}{Deqing Sun},
  \bibinfo{person}{Sofien Bouaziz}, \bibinfo{person}{Sean Fanello},
  \bibinfo{person}{Ping Tan}, {and} \bibinfo{person}{Yinda Zhang}.}
  \bibinfo{year}{2021}\natexlab{}.
\newblock \showarticletitle{HumanGPS: Geodesic PreServing Feature for Dense
  Human Correspondences}. In \bibinfo{booktitle}{\emph{Proceedings of the
  IEEE/CVF Conference on Computer Vision and Pattern Recognition}}.
  \bibinfo{pages}{1820--1830}.
\newblock


\bibitem[Vaswani et~al\mbox{.}(2017)]%
        {Transformer}
\bibfield{author}{\bibinfo{person}{Ashish Vaswani}, \bibinfo{person}{Noam
  Shazeer}, \bibinfo{person}{Niki Parmar}, \bibinfo{person}{Jakob Uszkoreit},
  \bibinfo{person}{Llion Jones}, \bibinfo{person}{Aidan N.~Gomez},
  \bibinfo{person}{Łukasz Kaiser}, {and} \bibinfo{person}{Illia Polosukhin}.}
  \bibinfo{year}{2017}\natexlab{}.
\newblock \showarticletitle{Attention is All You Need}. In
  \bibinfo{booktitle}{\emph{Proceedings of Neural Information Processing
  Systems (NeurlPS)}}. \bibinfo{pages}{5998--6008}.
\newblock


\bibitem[Wan et~al\mbox{.}(2020)]%
        {VCClothes}
\bibfield{author}{\bibinfo{person}{Fangbin Wan}, \bibinfo{person}{Yang Wu},
  \bibinfo{person}{Xuelin Qian}, \bibinfo{person}{Yixiong Chen}, {and}
  \bibinfo{person}{Yanwei Fu}.} \bibinfo{year}{2020}\natexlab{}.
\newblock \showarticletitle{When Person Re-Identification Meets Changing
  Clothes}. In \bibinfo{booktitle}{\emph{Proceedings of the IEEE/CVF Conference
  on Computer Vision and Pattern Recognition Workshops}}.
  \bibinfo{pages}{830--831}.
\newblock


\bibitem[Wang et~al\mbox{.}(2018)]%
        {MGN}
\bibfield{author}{\bibinfo{person}{Guanshuo Wang}, \bibinfo{person}{Yufeng
  Yuan}, \bibinfo{person}{Xiong Chen}, \bibinfo{person}{Jiwei Li}, {and}
  \bibinfo{person}{Xi Zhou}.} \bibinfo{year}{2018}\natexlab{}.
\newblock \showarticletitle{Learning Discriminative Features with Multiple
  Granularities for Person Re-Identification}. In
  \bibinfo{booktitle}{\emph{Proceedings of the 26th ACM international
  conference on Multimedia}}. \bibinfo{pages}{274–282}.
\newblock


\bibitem[Wang et~al\mbox{.}(2020)]%
        {NKUP}
\bibfield{author}{\bibinfo{person}{Kai Wang}, \bibinfo{person}{Zhi Ma},
  \bibinfo{person}{Shiyan Chen}, \bibinfo{person}{Jinni Yang},
  \bibinfo{person}{Keke Zhou}, {and} \bibinfo{person}{Tao Li}.}
  \bibinfo{year}{2020}\natexlab{}.
\newblock \showarticletitle{A Benchmark for Clothes Variation in Person
  Re-Identification}.
\newblock \bibinfo{journal}{\emph{IEEE Transactions on Pattern Analysis and
  Machine Intelligence}} \bibinfo{volume}{35}, \bibinfo{number}{12}
  (\bibinfo{year}{2020}), \bibinfo{pages}{1881--1898}.
\newblock


\bibitem[Wang et~al\mbox{.}(2022)]%
        {CoAttCC}
\bibfield{author}{\bibinfo{person}{Qizao Wang}, \bibinfo{person}{Xuelin Qian},
  \bibinfo{person}{Yanwei Fu}, {and} \bibinfo{person}{Xiangyang Xue}.}
  \bibinfo{year}{2022}\natexlab{}.
\newblock \showarticletitle{Co-Attention Aligned Mutual Cross-Attention for
  Cloth-Changing Person Re-Identification}. In
  \bibinfo{booktitle}{\emph{Proceedings of the Asian Conference on Computer
  Vision (ACCV)}}. \bibinfo{pages}{2270--2288}.
\newblock


\bibitem[Wei et~al\mbox{.}(2021)]%
        {MSMT172017}
\bibfield{author}{\bibinfo{person}{Longhui Wei}, \bibinfo{person}{Shiliang
  Zhang}, \bibinfo{person}{Wen Gao}, {and} \bibinfo{person}{Qi Tian}.}
  \bibinfo{year}{2021}\natexlab{}.
\newblock \showarticletitle{Person Transfer GAN to Bridge Domain Gap for Person
  Re-Identification}. In \bibinfo{booktitle}{\emph{Proceedings of the IEEE/CVF
  Conference on Computer Vision and Pattern Recognition}}.
  \bibinfo{pages}{79--88}.
\newblock


\bibitem[Wu et~al\mbox{.}(2021)]%
        {CVT}
\bibfield{author}{\bibinfo{person}{Haiping Wu}, \bibinfo{person}{Bin Xiao},
  \bibinfo{person}{Noel Codella}, \bibinfo{person}{Mengchen Liu},
  \bibinfo{person}{Xiyang Dai}, \bibinfo{person}{Lu Yuan}, {and}
  \bibinfo{person}{Lei Zhang}.} \bibinfo{year}{2021}\natexlab{}.
\newblock \showarticletitle{CvT: Introducing Convolutions to Vision
  Transformers}. In \bibinfo{booktitle}{\emph{Proceedings of the IEEE/CVF
  Conference on Computer Vision and Pattern Recognition}}.
  \bibinfo{pages}{22--31}.
\newblock


\bibitem[Xian et~al\mbox{.}(2023)]%
        {WACV2023}
\bibfield{author}{\bibinfo{person}{Yuqiao Xian}, \bibinfo{person}{Jinrui Yang},
  \bibinfo{person}{Fufu Yu}, \bibinfo{person}{Jun Zhang}, {and}
  \bibinfo{person}{Xing Sun}.} \bibinfo{year}{2023}\natexlab{}.
\newblock \showarticletitle{Graph-Based Self-Learning for Robust Person
  Re-Identification}. In \bibinfo{booktitle}{\emph{IEEE Workshop on
  Applications of Computer Vision (WACV)}}. \bibinfo{pages}{4789--4798}.
\newblock


\bibitem[Xu et~al\mbox{.}(2021)]%
        {Adversarial2021}
\bibfield{author}{\bibinfo{person}{Wanlu Xu}, \bibinfo{person}{Hong Liu},
  \bibinfo{person}{Wei Shi}, \bibinfo{person}{Ziling Miao}, {and}
  \bibinfo{person}{Feihu Chen}.} \bibinfo{year}{2021}\natexlab{}.
\newblock \showarticletitle{Adversarial Feature Disentanglement for Long-Term
  Person Re-Identification}. In \bibinfo{booktitle}{\emph{Proceedings of the
  30th International Joint Conference on Artificial Intelligence}}.
  \bibinfo{pages}{1201--1207}.
\newblock


\bibitem[Yan et~al\mbox{.}(2022)]%
        {Weakening2022}
\bibfield{author}{\bibinfo{person}{Yuming Yan}, \bibinfo{person}{Huimin Yu},
  \bibinfo{person}{Shuzhao Li}, \bibinfo{person}{Zhaohui Lu},
  \bibinfo{person}{Jianfeng He}, \bibinfo{person}{Haozhuo Zhang}, {and}
  \bibinfo{person}{Runfa Wang}.} \bibinfo{year}{2022}\natexlab{}.
\newblock \showarticletitle{Weakening the Influence of Clothing: Universal
  Clothing Attribute Disentanglement for Person Re-Identification}. In
  \bibinfo{booktitle}{\emph{Proceedings of the 31th International Joint
  Conference on Artificial Intelligence}}. \bibinfo{pages}{1523--1529}.
\newblock


\bibitem[Yang et~al\mbox{.}(2021)]%
        {PRCC}
\bibfield{author}{\bibinfo{person}{Qize Yang}, \bibinfo{person}{Ancong Wu},
  {and} \bibinfo{person}{Wei-Shi Zheng}.} \bibinfo{year}{2021}\natexlab{}.
\newblock \showarticletitle{Person Re-Identification by Contour Sketch Under
  Moderate Clothing Change}.
\newblock \bibinfo{journal}{\emph{IEEE Trans. Pattern Anal. Mach. Intell}}
  \bibinfo{volume}{43}, \bibinfo{number}{6} (\bibinfo{year}{2021}),
  \bibinfo{pages}{2029--2046}.
\newblock


\bibitem[Yang et~al\mbox{.}(2023)]%
        {Win-Win}
\bibfield{author}{\bibinfo{person}{Zhengwei Yang}, \bibinfo{person}{Xian
  Zhong}, \bibinfo{person}{Zhun Zhong}, \bibinfo{person}{Hong Liu},
  \bibinfo{person}{Zheng Wang}, {and} \bibinfo{person}{Shin'Ichi Satoh}.}
  \bibinfo{year}{2023}\natexlab{}.
\newblock \showarticletitle{Win-Win by Competition: Auxiliary-Free
  Cloth-Changing Person Re-Identification}.
\newblock \bibinfo{journal}{\emph{IEEE Transactions on Image Processing}}
  \bibinfo{volume}{32} (\bibinfo{year}{2023}), \bibinfo{pages}{2985--2999}.
\newblock


\bibitem[Yu et~al\mbox{.}(2018)]%
        {GeneralPerson1}
\bibfield{author}{\bibinfo{person}{Hong-Xing Yu}, \bibinfo{person}{Wu Ancong},
  {and} \bibinfo{person}{Zheng Wei-Shi}.} \bibinfo{year}{2018}\natexlab{}.
\newblock \showarticletitle{Unsupervised Person Re-Identification by Deep
  Asymmetric Metric Embedding}.
\newblock \bibinfo{journal}{\emph{IEEE Transactions on Pattern Analysis and
  Machine Intelligence}} \bibinfo{volume}{42}, \bibinfo{number}{4}
  (\bibinfo{year}{2018}), \bibinfo{pages}{956--973}.
\newblock


\bibitem[Yu et~al\mbox{.}(2020)]%
        {COCAS}
\bibfield{author}{\bibinfo{person}{Shijie Yu}, \bibinfo{person}{Shihua Li},
  \bibinfo{person}{Dapeng Chen}, \bibinfo{person}{Rui Zhao},
  \bibinfo{person}{Junjie Yan}, {and} \bibinfo{person}{Yu Qiao}.}
  \bibinfo{year}{2020}\natexlab{}.
\newblock \showarticletitle{COCAS: A Large-Scale Clothes Changing Person
  Dataset for Re-Identification}. In \bibinfo{booktitle}{\emph{Proceedings of
  the IEEE/CVF Conference on Computer Vision and Pattern Recognition}}.
  \bibinfo{pages}{3400--3409}.
\newblock


\bibitem[Zhang et~al\mbox{.}(2023)]%
        {SpeRe-Ranking}
\bibfield{author}{\bibinfo{person}{Renjie Zhang}, \bibinfo{person}{Yu Fang},
  \bibinfo{person}{Huaxin Song}, \bibinfo{person}{Fangbin Wan},
  \bibinfo{person}{Yanwei Fu}, \bibinfo{person}{Hirokazu Kato}, {and}
  \bibinfo{person}{Yang Wu}.} \bibinfo{year}{2023}\natexlab{}.
\newblock \showarticletitle{Specialized Re-Ranking: A Novel
  Retrieval-Verification Framework for Cloth Changing Person
  Re-Identification}.
\newblock \bibinfo{journal}{\emph{Pattern Recognition}}  \bibinfo{volume}{134}
  (\bibinfo{year}{2023}).
\newblock


\bibitem[Zhang et~al\mbox{.}(2019)]%
        {Densely2019}
\bibfield{author}{\bibinfo{person}{Zhizheng Zhang}, \bibinfo{person}{Cuiling
  Lan}, \bibinfo{person}{Wenjun Zeng}, {and} \bibinfo{person}{Zhibo Chen}.}
  \bibinfo{year}{2019}\natexlab{}.
\newblock \showarticletitle{Densely Semantically Aligned Person
  Re-Identification}. In \bibinfo{booktitle}{\emph{Proceedings of the IEEE/CVF
  Conference on Computer Vision and Pattern Recognition}}.
  \bibinfo{pages}{667--676}.
\newblock


\bibitem[Zheng et~al\mbox{.}(2015)]%
        {Market2015}
\bibfield{author}{\bibinfo{person}{Liang Zheng}, \bibinfo{person}{Liyue Shen},
  \bibinfo{person}{Lu Tian}, \bibinfo{person}{Shengjin Wang},
  \bibinfo{person}{Jingdong Wang}, {and} \bibinfo{person}{Qi Tian}.}
  \bibinfo{year}{2015}\natexlab{}.
\newblock \showarticletitle{Scalable Person Re-Identification: A Benchmark}. In
  \bibinfo{booktitle}{\emph{Proceedings of the IEEE/CVF International
  Conference on Computer Vision}}. \bibinfo{pages}{1116--1124}.
\newblock


\bibitem[Zheng et~al\mbox{.}(2019)]%
        {General5}
\bibfield{author}{\bibinfo{person}{Zhedong Zheng}, \bibinfo{person}{Liang
  Zheng}, {and} \bibinfo{person}{Yi Yang}.} \bibinfo{year}{2019}\natexlab{}.
\newblock \showarticletitle{Pedestrian Alignment Network for Large-scale Person
  Re-identification}.
\newblock \bibinfo{journal}{\emph{IEEE Transactions on Circuits and Systems for
  Video Technology}} \bibinfo{volume}{29}, \bibinfo{number}{10}
  (\bibinfo{year}{2019}), \bibinfo{pages}{3037--3045}.
\newblock


\end{thebibliography}

\end{document}